\documentclass[runningheads]{llncs}

\usepackage[arxiv]{eccv}

\usepackage{eccvabbrv}

\usepackage{graphicx}
\usepackage{booktabs}
\usepackage{tabularx}
\usepackage{array}

\usepackage[accsupp]{axessibility}  

\usepackage[pagebackref,breaklinks,colorlinks,citecolor=eccvblue]{hyperref}

\usepackage{orcidlink}

\usepackage{multirow}
\usepackage[normalem]{ulem}
\useunder{\uline}{\ul}{}
\usepackage{tcolorbox}
\tcbuselibrary{skins}
\usepackage{xcolor}

\newcommand{\method}{{\color[RGB]{0,0,0}{SVOR}}\xspace}
\newcommand{\methodfull}{{\color[RGB]{0,0,0}{Stable Video Object Removal}}\xspace}
\newcommand{\methodbf}{{\color[RGB]{0,0,0}\textbf{{SVOR}}}\xspace}

\begin{document}

\title{From Ideal to Real: Stable Video Object Removal under Imperfect Conditions}

\titlerunning{Stable Video Object Removal under Imperfect Conditions}

\author{
  Jiagao Hu$^{*}$ \and
  Yuxuan Chen$^{*}$ \and
  Fuhao Li$^{*}$ \and
  Zepeng Wang \and
  Fei Wang \and \\
  Daiguo Zhou \and
  Jian Luan
  }

\authorrunning{J.~Hu et al.}

\institute{
MiLM Plus, Xiaomi Inc. \\
\email{\{hujiagao,chenyuxuan7,lifuhao5\}@xiaomi.com}
}

\maketitle

\renewcommand{\thefootnote}{\fnsymbol{footnote}}
\footnotetext[1]{Equal contribution.}

\begin{abstract}
  Removing objects from videos remains difficult in the presence of real-world imperfections such as shadows, abrupt motion, and defective masks.
  Existing diffusion-based video inpainting models often struggle to maintain temporal stability and visual consistency under these challenges. We propose \textbf{Stable Video Object Removal (SVOR)}, a robust framework that achieves shadow-free, flicker-free, and mask-defect-tolerant removal through three key designs: (1) \textbf{Mask Union for Stable Erasure (MUSE)}, a windowed union strategy applied during temporal mask downsampling to preserve all target regions observed within each window, effectively handling abrupt motion and reducing missed removals; (2) \textbf{Denoising-Aware Segmentation (DA-Seg)}, a lightweight segmentation head on a decoupled side branch equipped with {Denoising-Aware AdaLN } and trained with mask degradation to provide an internal diffusion-aware localization prior without affecting content generation; and (3) \textbf{Curriculum Two-Stage Training}: where Stage I performs self-supervised pretraining on unpaired real-background videos with online random masks to learn realistic background and temporal priors, and Stage II refines on synthetic pairs using mask degradation and side-effect-weighted losses, jointly removing objects and their associated shadows/reflections while improving cross-domain robustness.
  Extensive experiments show that SVOR attains new state-of-the-art results across multiple datasets and degraded-mask benchmarks, advancing video object removal from ideal settings toward real-world applications.
  Project page: \url{https://xiaomi-research.github.io/svor/}
\end{abstract}

\begin{figure*}[t]
  \centering
  \includegraphics[width=\linewidth]{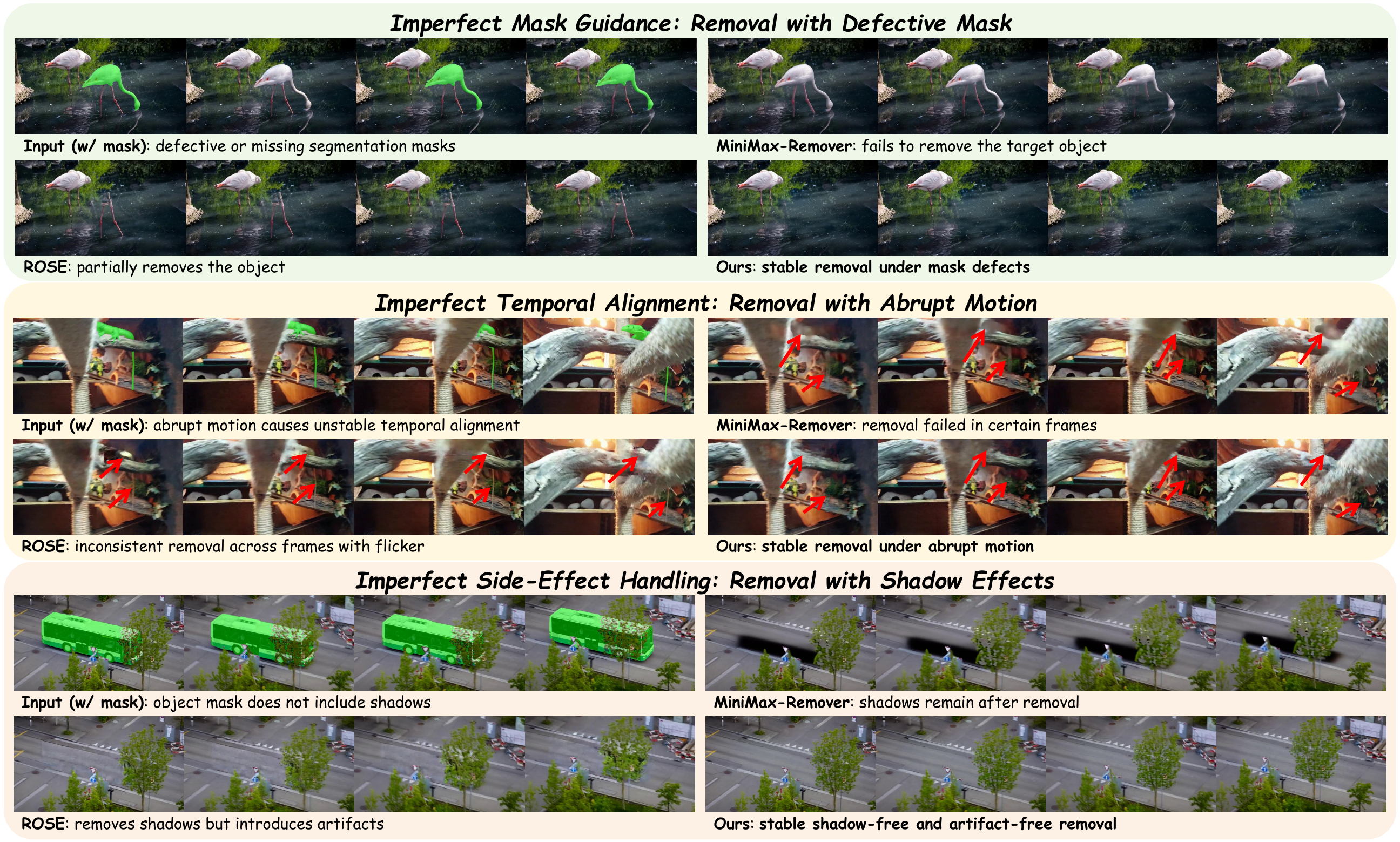}
  \caption{Results of our \methodfull compared with MiniMax-Remover~\cite{zi2025minimax} and ROSE~\cite{miao2025rose} in three common real-world challenges. The proposed \methodbf achieves stable and artifact-free removal.}
  \label{fig:teaser}
\end{figure*}

\section{Introduction}
\label{sec:intro}
Video object removal (VOR) aims to eliminate specified objects while reconstructing backgrounds that remain spatiotemporally consistent, and is widely used in video editing, post-production, and AR. Recent VOR approaches have made significant progress in aspects such as inference efficiency~\cite{zi2025minimax, litman2026editctrl} and side-effect suppression (\eg, shadows and reflections)~\cite{lee2025generative, miao2025rose, kushwaha2026object}, achieving impressive results. However, their performance often degrades under realistic conditions—such as abrupt object motion, imperfect masks, and real-world side effects. The root cause lies in the overly idealized deployment assumptions and the underexplored flaws of existing inpainting pipelines.

Mainstream methods typically assume high-quality segmentation masks as guidance, \ie, (i) a mask on every frame and (ii) sufficiently fine boundaries. In real scenarios these assumptions break: even advanced segmenters (\eg, SAM series~\cite{kirillov2023segment, ravi2024sam, carion2025sam}) can suffer target loss, weakened ID consistency, or mis-segmentation under occlusion, fast motion, appearance ambiguities, and fine structures (\eg, hair), producing inaccurate or missing masks and thus artifacts/residues in the final results. Moreover, human annotations are inherently sparse; asking users to inspect and correct masks frame-by-frame is impractical.

Beyond annotation noise, preprocessing and training further amplify instability. Before feeding to the backbone, masks are commonly temporally compressed~\cite{zi2025minimax} or downsampled~\cite{lee2025generative, jiang2025vace}, inevitably losing temporal localization. With rapid motion or missing frames, location cues within a compression window are attenuated or swallowed, breaking alignment and causing missed removals and flicker. On the training side, synthetic paired data help suppress side effects such as shadows/reflections~\cite{lee2025generative, miao2025rose}, but relying on synthetic-only supervision incurs significant domain shift, which still produces artifacts on real videos.

Altogether, these vulnerabilities form a conceptual stability taxonomy for VOR, spanning annotation, preprocessing, and training. To systematically tackle these dimensions, we propose \textbf{Stable Video Object Removal (SVOR)}, enabling stable, high-quality removal under three types of imperfection, as shown in~\cref{fig:teaser}.

\textbf{Imperfect Mask Guidance.} During training, we explicitly apply mask degradation (temporal sparsity + spatial degradation) to encourage removal under imperfect external cues. To suppress false removal induced by degraded masks, we introduce a lightweight \textit{Denoising-Aware Segmentation} head (\textit{DA-Seg}) with \textit{Denoising-Aware AdaLN} (\textit{DA-AdaLN}). Attached to an auxiliary branch, this denoising-aware head provides diffusion-specific localization priors to complement degraded masks. Unlike inserting a mask head into the backbone~\cite{miao2025rose} or feeding predicted masks back into it~\cite{zheng2023ciri}, we decouple localization from generation. DA-Seg is trained for localization only and never conditions backbone denoising, thereby preserving content synthesis and stabilizing erasure under defective-mask guidance.

\textbf{Imperfect Temporal Alignment.} We introduce \textit{Mask Union for Stable Erasure (MUSE)}, an effective remedy for abrupt motion under temporal downsampling. For each compression window, MUSE retains the union of all mask locations observed in the window, preserving short-lived object positions that would otherwise be dropped. This maximizes coverage without extra parameters and substantially reduces under-erasure and ghosting in abrupt motion frames, while our experiments show that MUSE has negligible impact on non-erased regions. \textbf{To our knowledge, we are the first to report that mask downsampling under abrupt motion leads to systematic under-erasure in widely-used pipelines.}

\textbf{Imperfect Side-Effect Handling.} We adopt a \textit{Curriculum Two-Stage Training} strategy that decouples ``removing the object and side effects'' from ``restoring real backgrounds.'' Stage I pretrains on unpaired real background videos with online random masks, fostering a background-first reconstruction prior. Stage II refines with synthetic paired supervision under mask degradation and applies weighted supervision to side-effect regions, strengthening shadow/reflection cleanup. The two stages act synergistically to reduce optimization difficulty and substantially improve shadow removal quality.

\textbf{Stable Video Object Removal is in the details. Our contributions are threefold:}

\begin{itemize}
    \item We identify a failure mode where \textit{temporal mask downsampling misses targets under abrupt motion}, and propose \textbf{MUSE} to preserve short-lived object locations and reduce under-erasure, ghosting, and flicker.
    \item We propose a lightweight decoupled side-branch segmentation head \textbf{DA-Seg} to provide a stable internal localization prior under defective masks.
    \item We build a \textbf{S}tability-centric framework for \textbf{VOR} under imperfect conditions, using two-stage training to improve temporal stability, mask robustness, and side-effect suppression.
\end{itemize}

We further introduce RORD-50, a paired real-world test set for video object removal, based on RORD~\cite{sagong2022rord}. Across DAVIS~\cite{pont20172017}, ROSE Bench~\cite{miao2025rose}, RORD-50, and the corresponding degraded-mask variants, our method consistently outperforms prior art.

\section{Related Works}
\label{sec:related}

\subsection{Non-diffusion methods}
Non-diffusion methods propagate known pixels across frames using 3D CNNs~\cite{hu2020proposal,wang2019video,chan2022basicvsr++}, optical flow~\cite{kim2019deep,zhang2022inertia,li2022towards}, or homography~\cite{lin2019tsm,zou2021progressive,ke2021occlusion}, but suffer from limited temporal context and alignment errors. Transformer-based approaches ~\cite{zhang2022flow,zeng2020learning,liu2021fuseformer,ren2022dlformer} improve long-range coherence via spatio-temporal attention, and flow-based methods~\cite{zhou2023propainter} combines flow completion, dual-domain warping, or sparse attention for robust propagation. However, these methods struggle with large masks or occlusions, often producing structural ambiguity, texture loss, or flickering.

\subsection{Diffusion-based methods}
Diffusion models and video transformers now dominate video object removal, with methods guided by (i) \emph{mask-based} inpainting~\cite{zi2025minimax,li2025diffueraser,miao2025rose,lee2025generative, kushwaha2026object} and (ii) \emph{text-guided}~\cite{zhang2024avid,zi2025cococo,hu2024vivid,yang2025mtv,bian2025videopainter,wang2023videocomposer, litman2026editctrl} local edits that preserve temporal coherence. Motion/structure guidance with trainable temporal attention improves consistency and control (AVID~\cite{zhang2024avid}, CoCoCo~\cite{zi2025cococo}). VideoPainter~\cite{bian2025videopainter} decouples foreground synthesis from background preservation with a plug-and-play context encoder and ID resampling. For pure removal, many drop text to avoid semantic drift: DiffuEraser~\cite{li2025diffueraser} enlarges the temporal receptive field via prior-frame initialization; MiniMax-Remover~\cite{zi2025minimax} uses minimax noise training for fast, high-quality removal. Beyond content, ROSE~\cite{miao2025rose}, Generative Omnimatte~\cite{lee2025generative} and Object-WIPER~\cite{kushwaha2026object} address side-effect disambiguation (\eg, shadows, reflections). Unlike prior work, we target stable removal of targets and their effects under imperfect mask guidance.

\begin{figure*}[t]
    \centering
    \includegraphics[width=\textwidth]{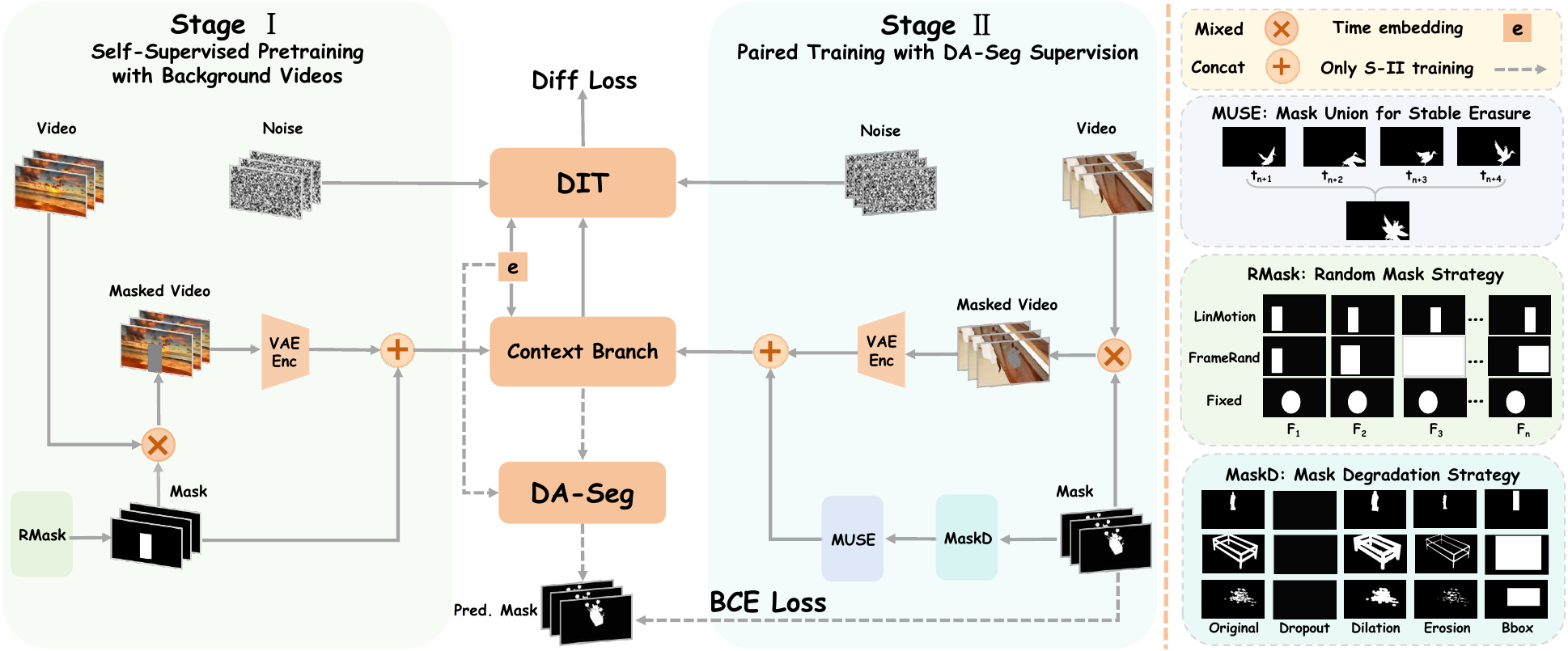}
    \caption{\textbf{The framework of \method. }Stage I: pretrain on unpaired real-world background videos using Random Mask Strategy to simulate object motion. Stage II: refine on paired synthetic data with Mask Degradation to mimic imperfect masks, where DA-Seg complements defective guidance. MUSE performs windowed union retention during mask temporal downsampling, preventing loss of dynamic location information.}
    \label{fig:framework}
\end{figure*}

\section{Method}
\label{sec:method}
\subsection{Architectural Overall}

Following the side-branch conditioning design of VACE~\cite{jiang2025vace}, we inject the input frames and masks into the backbone DiT~\cite{peebles2023scalable} through a lightweight context branch, enabling mask-guided video generation without interfering with the main denoising stream. Built upon this architecture, we propose a curriculum-style two-stage training framework, as illustrated in \cref{fig:framework}, to progressively address the challenges of video object removal in real-world scenarios.
In \textbf{Stage~I}, the model is self-supervised on unpaired real-world background videos to prime the base removal capability. In \textbf{Stage~II}, the model is trained on paired synthetic data to remove shadows and other side effects, while improving removal stability under low-quality masks.

For more stable results under imperfect masks, we introduce a \textit{Denoising-Aware Segmentation} head (\textit{DA-Seg}) to provide diffusion-specific localization priors that complement defective masks.
To mitigate missed removals caused by abrupt motion frames, we revise the temporal mask downsampling by incorporating \textit{Mask Union for Stable Erasure (MUSE)}.

In a word, Stage~I learns a strong background completion prior, while Stage~II improves robustness to imperfect masks via DA-Seg supervision, and MUSE further corrects structural misalignment introduced by temporal mask compression under abrupt motion.

\subsection{Stage I: Self-Supervised Pretraining with Background Videos}
\label{sec:stage1}

Existing video inpainting models exhibit some ``removal'' capability, but when directly applied to object removal they often suffer from \emph{undesired regeneration}, \ie, re-synthesizing foreground-like content inside masked regions. While finetuning with paired data can mitigate this issue~\cite{miao2025rose}, such data are difficult to obtain. Prior works~\cite{zi2025minimax, lee2025generative} therefore rely on copy-paste to synthesize pseudo pairs; however, copy-paste can introduce both \emph{physical} inconsistencies (mismatched object--shadow/reflection relations) and \emph{semantic} inconsistencies (objects that do not fit the scene context). As a result, the model may learn spurious cues for side effects, which can hinder subsequent effect removal.

We instead propose a \textbf{background-only self-supervised pretraining} stage that requires no explicit paired data. Using videos without salient foreground objects, we apply online \textbf{random masks} and optimize the model to reconstruct the missing regions. Importantly, this stage does not enforce any (potentially wrong) object--side-effect correspondence; instead, it learns a \emph{context-consistent} background completion and base-erasure prior that discourages foreground-like synthesis under occlusion. This provides a strong initialization for Stage~II, where the model can focus on learning side-effect suppression with paired supervision (see \cref{sec:exp_ablation,sec_supp:bg_train}).

\paragraph{\textbf{Background Data Construction.}}
\label{sec:bg_data_construction}
We mine background-only clips from publicly available datasets~\cite{nan2024openvid, li2023videogen, xue2025ultravideo, stergiou2022adapool, wang2025wisa} using a multi-stage filtering pipeline, including quality filtering, VLM-based scene selection, and open-world detection/segmentation. Clips containing salient foreground regions ($>30\%$ of frames) are discarded, followed by lightweight manual verification. This process yields approximately 49K diverse background videos spanning diverse scenes, viewpoints, and lighting conditions (details in \cref{sec_supp:bgdata} in the supplementary material).

\paragraph{\textbf{Random Mask.}}
\label{sec:random_mask}
In Stage~I, mask semantics are intentionally deemphasized; instead, we focus on diverse \emph{spatiotemporal occlusion patterns}. We adopt an online random mask strategy that composes simple spatial shapes (rectangles, ellipses, full-frame masks) with varied temporal dynamics, including static, intermittent, jittered, and trajectory-based motion. Masks are resampled on-the-fly for each clip, encouraging generalization across occlusion locations, durations, and motions.
This design explicitly biases the model toward background reconstruction rather than object synthesis, and lays a strong foundation for stable erasure under imperfect masks in Stage~II. (details in \cref{sec_supp:Rmask}).

\paragraph{\textbf{Self-Supervised Pretraining Objective.}}
\label{sec:stage1_opt}

Stage~I follows the standard diffusion noise prediction paradigm and is optimized in a fully self-supervised manner. Given a background video $x$ and a randomly generated mask $m$, we construct the occluded input $\tilde{x}=\operatorname{masked}(x,m)$ and train the model to recover the original content in the masked regions.

Formally, let $\mathrm{VAE}(x)=\mathbf{z}_0$, and obtain the noisy latent via forward diffusion:
\begin{equation}
    {z}_t = \alpha_t {z}_0 + \sigma_t {\epsilon}, \quad
    {\epsilon} \sim \mathcal{N}(0, I), \quad
    t \sim \{1, \dots, T\}.
\end{equation}
Here, ${\epsilon}$ represents the noise added to the latent, and $t$ is the time step in the diffusion process.
The diffusion loss is defined as
\begin{equation}
\mathcal{L}_{\text{diff}}
= \mathbb{E}_{x,m,{\epsilon},t}\!\left[
  \left\|{\epsilon}-{\epsilon_{\theta}}\!\left({z}_t,\,t,\,\tilde{x}\right)\right\|_2^{2}
\right].
\end{equation}
${\epsilon}_{\theta}({z}_t, t, \tilde{x})$ denotes the noise predicted by the denoising network with parameters $\theta$ at diffusion step $t$ for the noisy latent ${z}_t$, conditioned on the occluded input $\tilde{x}$.

\subsection{Stage II: Paired Training with DA-Seg Supervision}
\label{sec:stage2}

Building upon the strong background completion prior learned in Stage~I, Stage~II fine-tunes the model on paired synthetic data to achieve clean and stable object removal under imperfect mask guidance. In real-world scenarios, removal masks are often temporally sparse, spatially inaccurate, or partially missing, which can lead to residual artifacts and inconsistent erasure. To address these challenges, Stage~II introduces three complementary components: (i) \textbf{mask degradation} to improve robustness under noisy supervision, (ii) a \textbf{Denoising-Aware Segmentation (DA-Seg)} head to provide a stable internal localization prior, and (iii) \textbf{Mask Union for Stable Erasure (MUSE)} to correct structural misalignment introduced by temporal mask compression. Together, these designs enable reliable removal under weak and imperfect masks.

\paragraph{\textbf{Mask Degradation.}}
\label{sec:mask_degradation}

High-quality, pixel-accurate masks are rarely available in practice due to motion blur, occlusion, lighting variation, and annotation noise. To improve robustness to such imperfect supervision, we apply \emph{mask degradation} during Stage~II training, encouraging the model to perform clean erasure under weak and noisy guidance.

Concretely, starting from the accurate ground-truth mask $\mathbf{M}$, we online sample degraded variants and random mixtures, including: (i) frame-level dropout (20\%--99\%), (ii) morphological erosion and dilation to perturb boundaries, and (iii) coarse localizers (bbox fits) to mimic temporally sparse or missing masks. These degradations are randomly composed per training sample, forcing the model to rely less on precise mask boundaries and more on contextual and temporal cues, thereby improving robustness to real-world mask imperfections.

\paragraph{\textbf{Denoising-Aware Segmentation (DA-Seg).}}
\label{sec:da_seg}

Under defective mask guidance, accurate localization of the erasure object remains critical. We therefore introduce a lightweight side-branch segmentation head, termed \emph{DA-Seg}, which provides an internal localization prior while remaining fully decoupled from the backbone DiT. This design allows the model to focus on the target region without perturbing the backbone’s generative states.

Formally, given the side-branch features $f \in \mathbb{R}^{B \times L \times C}$, DA-Seg predicts a soft mask via a \emph{Denoising-Aware AdaLN} (DA-AdaLN) followed by an MLP:
\begin{equation}
s = \mathrm{MLP}\!\left(\text{DA-AdaLN}(f)\right).
\end{equation}
After unpatchifying, $s$ is mapped to $\hat{M} \in [0,1]^{B \times F_p \times H_p \times W_p}$, representing the model’s internal estimate of the \emph{to-be-erased} region.

DA-AdaLN conditions the segmentation head on the diffusion timestep embedding $e \in \mathbb{R}^{B \times C}$, following the adaptive normalization design in DiT~\cite{peebles2023scalable}. Specifically, the shift and scale parameters $\beta, \gamma \in \mathbb{R}^{B \times C}$ are produced by learnable modulation parameters $m \in \mathbb{R}^{1 \times 2 \times C}$:
\begin{equation}
(\beta, \gamma) = \mathrm{chunk}\!\left(m + \mathrm{unsqueeze}(e, 1),\, 2\right).
\end{equation}
This denoising-aware conditioning enables the segmentation head to adapt across noise levels in a coarse-to-fine manner, improving stability under high-noise diffusion steps. We empirically verify the necessity of timestep conditioning by a dedicated ablation in the supplementary material (\cref{sec_supp:seghead}), where DA-AdaLN consistently outperforms standard LayerNorm under defective-mask supervision.

Importantly, the predicted mask is not fed back into the backbone. It is used exclusively for supervision against the downsampled ground-truth mask $M_{\text{gt}}^{\downarrow}$, following a \emph{side-branch localization, backbone generation} paradigm that preserves generative capacity while stabilizing removal under imperfect masks.

\paragraph{\textbf{Mask Union for Stable Erasure (MUSE).}}
\label{sec:muse}

Despite mask degradation and DA-Seg supervision, we observe persistent failures under abrupt motion or temporally sparse masks. Our analysis shows that these failures stem from a structural misalignment introduced during temporal mask compression, rather than insufficient robustness.

Specifically, diffusion-based video editing frameworks typically compress masks along the temporal axis to match the latent resolution. For example, VACE~\cite{jiang2025vace} applies $4\times$ temporal downsampling using nearest-neighbor sampling. Under abrupt motion, this strategy selects a single frame per temporal window, leading to temporal truncation and displacement bias; if the selected frame is empty or weak, the compressed mask may collapse entirely. These issues manifest as missed removals and smearing artifacts. Similar failure modes are observed in several recent methods~\cite{lee2025generative, zi2025minimax, miao2025rose} (see \cref{fig:teaser} and \cref{sec_supp:muse_others}).

To address this issue, we propose \emph{Mask Union for Stable Erasure (MUSE)}, a simple yet effective fix applied during Stage~II training and inference. MUSE aligns with the VAE temporal compression scheme by adopting a \emph{first-frame anchoring} and \emph{grouped temporal union} strategy. Concretely, the first mask is mapped directly to the first compressed latent frame, while subsequent masks are grouped according to the temporal compression ratio (4 by default). For each group, we compute an element-wise union (\ie, a temporal logical $OR$) to produce the compressed mask. This preserves any location that appears within each window, preventing the loss of dynamic content while maintaining strict alignment with the latent sequence. Despite its simplicity, MUSE can be applied in a plug-and-play manner and consistently improves existing models, as shown in the supplementary material (\cref{sec_supp:muse_others}).

\paragraph{\textbf{Training Objective.}}
\label{sec:seg_loss}

DA-Seg is supervised using a binary cross-entropy (BCE) loss between its prediction $\hat{M}$ and the downsampled ground-truth mask $M_{\text{gt}}^{\downarrow}$:

\begin{equation}
\mathcal{L}_{\text{seg}} = \mathrm{BCE}\!\left(\hat{M},\, M_{\text{gt}}^{\downarrow}\right).
\end{equation}

To further reduce residual artifacts in side-effect regions such as shadows and reflections, we adopt a weighted diffusion loss based on the side-effect mask $D$, computed following ROSE~\cite{miao2025rose}:

\begin{equation}
\mathcal{L}_{\text{diff}} = \mathbb{E}_{x,M,\epsilon,t}\left[\sum_{q} w^q \left\|\epsilon_q - \epsilon_\theta^q(z_t, t, \tilde{x})\right\|_2^2\right],
\text{where } {w^q} = \begin{cases} \lambda_w, & D^q = 1, \\ 1, & \text{otherwise.} \end{cases}
\end{equation}

The final Stage~II objective is a weighted combination:

\begin{equation}
\mathcal{L}_{\text{total}}
= \mathcal{L}_{\text{diff}} + \lambda_s\,\mathcal{L}_{\text{seg}} .
\end{equation}

We find the method to be robust to moderate variations of $\lambda_w$ and $\lambda_s$, and use fixed values $\lambda_w = 2$ and $\lambda_s = 0.2$ for all experiments.
\section{Experiments}
\label{sec:exp}

\definecolor{myred}{RGB}{255,0,0}
\definecolor{myblue}{RGB}{72,116,203}
\definecolor{mygreen}{RGB}{48,192,180}
\definecolor{myyellow}{RGB}{255,184,130}
\definecolor{myorange}{RGB}{242,182,2}
\definecolor{mypurple}{RGB}{112,45,160}

\subsection{Experiment Settings}

\paragraph{\textbf{Training Dataset.}}
Our training follows the Curriculum Two-Stage Training described in \cref{sec:method}.
Stage I uses large-scale background videos collected from publicly available sources (details in \cref{sec_supp:bgdata} in the supplementary material) to learn real background and temporal priors.
Stage II finetunes on the ROSE dataset~\cite{miao2025rose}, which provides $\sim$16k paired triplets of \{\emph{original video}, \emph{object mask}, and \emph{groundtruth result}\}. This stage introduces mask degradation and DA-Seg to strengthen robustness under non-ideal masks.

\paragraph{\textbf{Evaluation Dataset.}}
We evaluate on three datasets, \ie
(1)~\textbf{DAVIS}~\cite{pont20172017} which contains 90 real-world videos without paired groundtruth. We use all these 90 videos together with corresponding all instances' masks for evaluations.
(2)~\textbf{ROSE Bench}~\cite{miao2025rose} which is a benchmark with 3 components. We choose the public available subset which consists of 60 synthetic video triplets with origin, mask and object-removed videos. It contains various objects and 6 types of side effects.
(3)~\textbf{RORD-50}: a new paired real-world benchmark constructed from RORD~\cite{sagong2022rord}. We replicate the background image into a video to form groundtruth, and select 50 pairs whose background area best aligns with the corresponding input (details in \cref{sec_supp:rord50}). Also, we manually segment the objects from each frames for the perfect masks.
This dataset bridges the domain gap between real and synthetic data while enabling paired evaluation.

\paragraph{\textbf{Evaluation Metrics.}}
Following previous image and video inpainting methods, we use PSNR~\cite{hore2010image} and SSIM~\cite{wang2004image} to quantitatively evaluate the generative quality. However, these metrics can only evaluate the reconstruction of non-removal regions for unpaired data (e.g., DAVIS), which reflects background consistency rather than removal quality. Thus we also adopt the ReMOVE~\cite{chandrasekar2024remove} score, which is a reference-free metric to assess removal performance by comparing the reconstruction in the target regions with the background regions. Also, we use the Temporal Flickering metric (denoted as TF) in VBench~\cite{huang2024vbench} to evaluate temporal consistency.
Additionally, we conduct an LLM-based perceptual evaluation using GPT-4o (denoted as GPT), which scores both removal correctness and visual plausibility (details in \cref{sec_supp:gpt_eval}).

\paragraph{\textbf{Implementation Details.}}
We implement the \method model based on Wan2.1-VACE-1.3B~\cite{jiang2025vace}, adding a segmentation head to the context branch.
We use 8 Nvidia H100 GPUs with a batch size of 3 and a learning rate of $1\times10^{-4}$ for 5 epochs per training stage.

\begin{table*}[t]
\caption{Quantitative comparison of different methods. The best performance is highlighted in \textbf{bold}, while the second-best is \underline{underlined}. All results are reproduced using their official implementations to ensure fairness. For gen-omni, we use the recommended CogVideoX-Fun-V1.5-5b-InP version.}
\label{tab:cmp}
\centering
\newcolumntype{I}{!{\vrule width 1.2pt}} 

\resizebox{\linewidth}{!}{%
\begin{tabular}{cIc|c|c|c|cIc|c|c|c|cIc|c|c|c|c}
\toprule
\multicolumn{1}{cI}{} & \multicolumn{5}{cI}{DAVIS} & \multicolumn{5}{cI}{ROSE Bench} & \multicolumn{5}{c}{RORD-50} \\
\cline{2-16}
\multicolumn{1}{cI}{} &
\text{bgPSNR↑} & \text{bgSSIM↑} & \text{TF↓} & \text{ReMove↑} & \text{GPT↑} &
\text{PSNR↑} & \text{SSIM↑} & \text{TF↓} & \text{ReMove↑} & \text{GPT↑} &
\text{PSNR↑} & \text{SSIM↑} & \text{TF↓} & \text{ReMove↑} & \text{GPT↑} \\
\midrule
FuseFormer~\cite{liu2021fuseformer}
& 29.51 & 0.8600 & 0.9921 & \underline{0.8763} & 9.379
& 25.75 & 0.8847 & 0.9921 & 0.9058 & 9.776
& 27.04 & 0.8576 & 0.9961 & 0.9167 & 10.79 \\

FGT~\cite{zhang2022flow}
& 30.74 & 0.9025 & 0.9537 & 0.8731 & 9.465
& 26.35 & 0.9059 & 0.9916 & 0.8927 & 10.89
& 27.51 & 0.8804 & 0.9961 & 0.9117 & 10.73 \\

Propainter~\cite{zhou2023propainter}
& {\bfseries{36.12}} & {\bfseries{0.9753}} & 0.9529 & 0.8607 & 9.446
& 25.14 & \underline{0.9241} & \textbf{0.9903} & 0.8204 & 7.780
& 29.54 & \underline{0.9367} & 0.9958 & 0.9040 & 10.16 \\

\midrule
DiffuEraser~\cite{li2025diffueraser}
& \underline{33.76} & \underline{0.9467} & 0.9522 & 0.8626 & 10.25
& 26.83 & 0.9040 & 0.9917 & 0.8837 & 10.72
& 29.84 & 0.9345 & 0.9958 & 0.9126 & 11.53 \\

VACE~\cite{jiang2025vace}
& 27.06 & 0.8656 & \textbf{0.9471} & 0.7081 & 5.263
& 22.71 & 0.8802 & \underline{0.9913} & 0.7154 & 7.617
& 19.21 & 0.8622 & \underline{0.9856} & 0.6842 & 3.962 \\

gen-omni~\cite{lee2025generative}
& 27.56 & 0.8586 & 0.9643 & 0.8742 & \underline{11.58}
& 27.08 & 0.8831 & 0.9925 & 0.8998 & 12.45
& 30.68 & 0.9159 & 0.9993 & \underline{0.9174} & \underline{13.36} \\

minimax~\cite{zi2025minimax}
& 30.00 & 0.8820 & 0.9555 & 0.8711 & 10.62
& 26.30 & 0.8950 & 0.9918 & 0.8960 & 11.23
& 28.85 & 0.9273 & 0.9973 & 0.9155 & 11.70 \\

ROSE~\cite{miao2025rose}
& 28.11 & 0.8748 & 0.9559 & 0.8683 & 10.48
& \underline{31.12} & 0.9170 & \underline{0.9913} & \underline{0.9081} & \underline{12.76}
& \underline{30.93} & 0.9186 & 0.9971 & \underline{0.9174} & 12.88 \\

\midrule
Ours
& 28.29 & 0.9092 & \underline{0.9510} & {\bfseries{0.8800}} & {\bfseries{12.34}}
& {\bfseries{31.47}} & {\bfseries{0.9335}} & \textbf{0.9903} & {\bfseries{0.9082}} & {\bfseries{13.18}}
& {\bfseries{31.26}} & {\bfseries{0.9378}} & \textbf{0.9851} & {\bfseries{0.9179}} & {\bfseries{13.82}} \\
\bottomrule
\end{tabular}
}
\end{table*}

\begin{table}[t]
\caption{User study results on DAVIS. Our \method achieves the highest overall perceptual performance. 
}
\label{tab:user_study}
\centering
\begin{tabular}{c|ccccc}
\toprule
 & Propainter~\cite{zhou2023propainter} & gen-omni~\cite{lee2025generative} & minimax~\cite{zi2025minimax} & ROSE~\cite{miao2025rose} & Ours \\
\midrule
Erasure & 0.897 & {\ul 0.904} & 0.926 & 0.934 & \textbf{0.978} \\
Completion & 0.228 & \textbf{0.588} & 0.412 & 0.353 & {\ul 0.551} \\
\midrule
Average & 0.563 & {\ul 0.746} & 0.669 & 0.644 & \textbf{0.765} \\
\bottomrule
\end{tabular}
\end{table}

\begin{figure*}[t]
    \centering
    \includegraphics[width=1\linewidth]{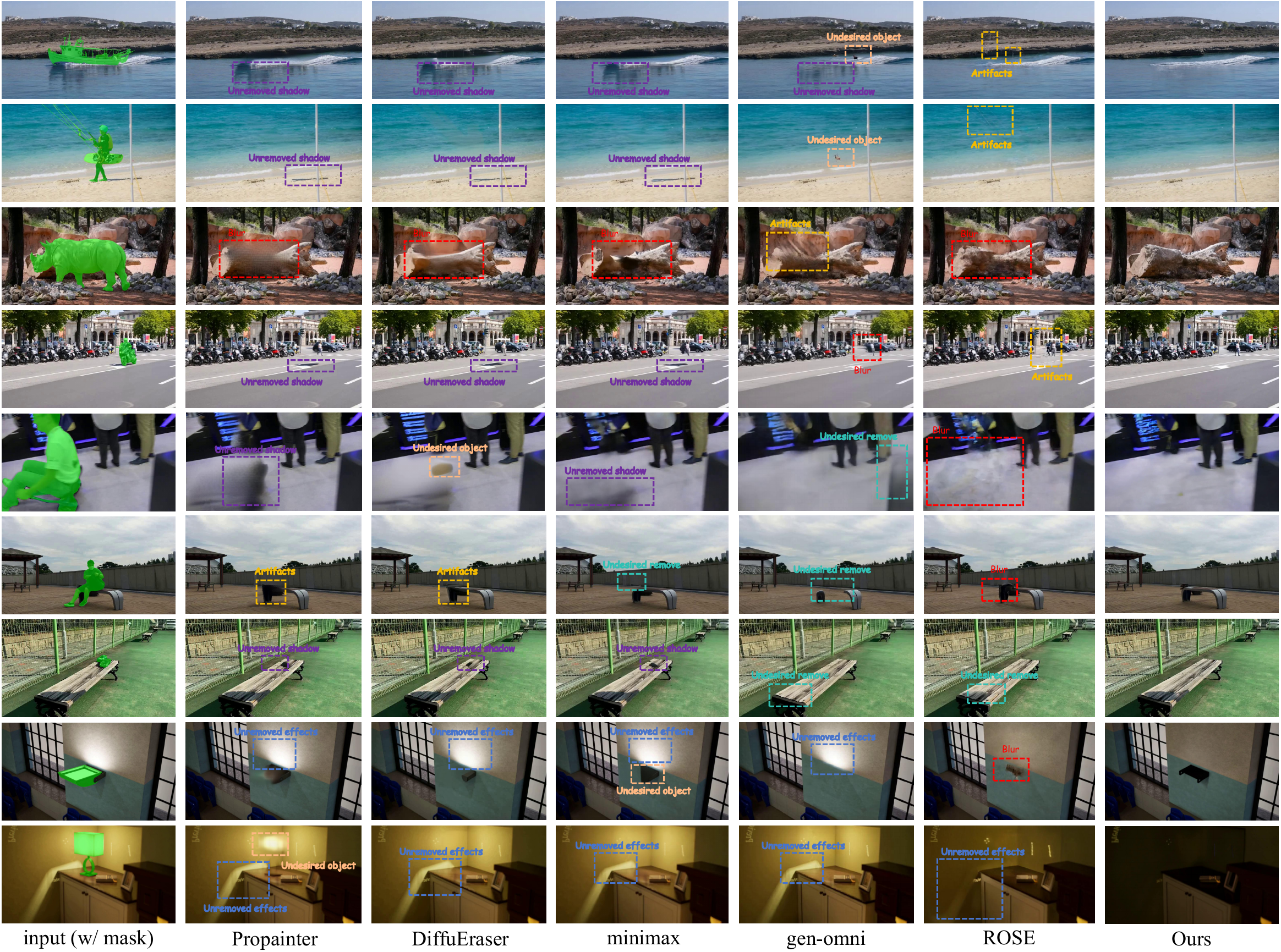}
    \caption{Qualitative comparison between our \method and several state-of-the-art methods on real-world and synthetic samples. Previous methods facing issues like \textcolor{myyellow}{Undesired object}, \textcolor{myorange}{Artifacts}, \textcolor{myred}{Blur}, \textcolor{mygreen}{Undesired remove}, \textcolor{mypurple}{Unremoved shadow}, \textcolor{myblue}{Unremoved effects}. Our \method achieves consistently cleaner removal, fewer artifacts, and better shadow handling.}
    \label{fig:qualitative_result}
\end{figure*}

\subsection{Quantitative Evaluation}

\paragraph{\textbf{Automatic Evaluation.}}
We compare our method on all three datasets with several state-of-the-art (SOTA) models, includes non-diffusion methods (\ie, FuseFormer~\cite{liu2021fuseformer}, FGT~\cite{zhang2022flow}, Propainter~\cite{zhou2023propainter}) and diffusion-based models (\ie, DiffuEraser~\cite{li2025diffueraser}, VACE~\cite{jiang2025vace}, gen-omni~\cite{lee2025generative}, minimax~\cite{zi2025minimax}, ROSE~\cite{miao2025rose}). The experimental results are shown in \cref{tab:cmp}. From the table, we can see that our method achieves the best ReMOVE and GPT scores on all three datasets. On the ROSE Bench and RORD-50 datasets, which have paired groundtruth, our method also outperforms other methods in PSNR and SSIM metrics.
It is noteworthy that since the DAVIS lacks paired groundtruth, its PSNR and SSIM scores are computed only on the non-removal regions (denoted as bgPSNR and bgSSIM), reflecting the impact on the background rather than the target removal. Given that our method, along with gen-omni~\cite{lee2025generative} and ROSE~\cite{miao2025rose}, can effectively remove shadows and other side effects, these inevitable background changes result in slightly lower scores for these metrics.

\paragraph{\textbf{User Study.}}
In addition, we conduct a user study on DAVIS videos. Each participant was shown the masked input alongside results from Propainter~\cite{zhou2023propainter}, gen-omni~\cite{lee2025generative}, minimax~\cite{zi2025minimax}, ROSE~\cite{miao2025rose}, and our \method. Participants scored each result from two perspectives: (1) \textit{Erasure}: whether the target object was fully removed, and (2) \textit{Completion}: whether the filled region is painted with noticeable flaws. Each criterion is scored as \{0, 0.5, 1\}, and the final score is the average across users.

We recruited 15 participants for this study. The results, summarized in \cref{tab:user_study}, demonstrate that our method achieves the highest score in the \textit{Erasure} dimension and ranks second in \textit{Completion}. When averaging the two, our method achieves the best overall performance, demonstrating a favorable balance between effective object removal and visually coherent background restoration.

\subsection{Qualitative Results}

\Cref{fig:qualitative_result} shows the object removal results of our method and several SOTA methods on real-world scenes from the DAVIS and RORD-50 videos, as well as synthetic ROSE Bench samples. The results clearly show several issues in existing methods, such as generating undesired objects, incorrectly erasing non-target objects, producing artifacts, blurring, and failing to remove shadows and other effects. In contrast, our method consistently removes the object and associated effects without artifacts or blurring. Despite being partially trained on synthetic ROSE data, our two-stage curriculum training strategy enables strong generalization to real-world dynamics, minimizing artifacts and blurring when comparied with ROSE~\cite{miao2025rose}.

\subsection{Stability of Removal}
\label{sec:exp_stablity}

\begin{figure}[t]
    \centering
    \includegraphics[width=\linewidth]{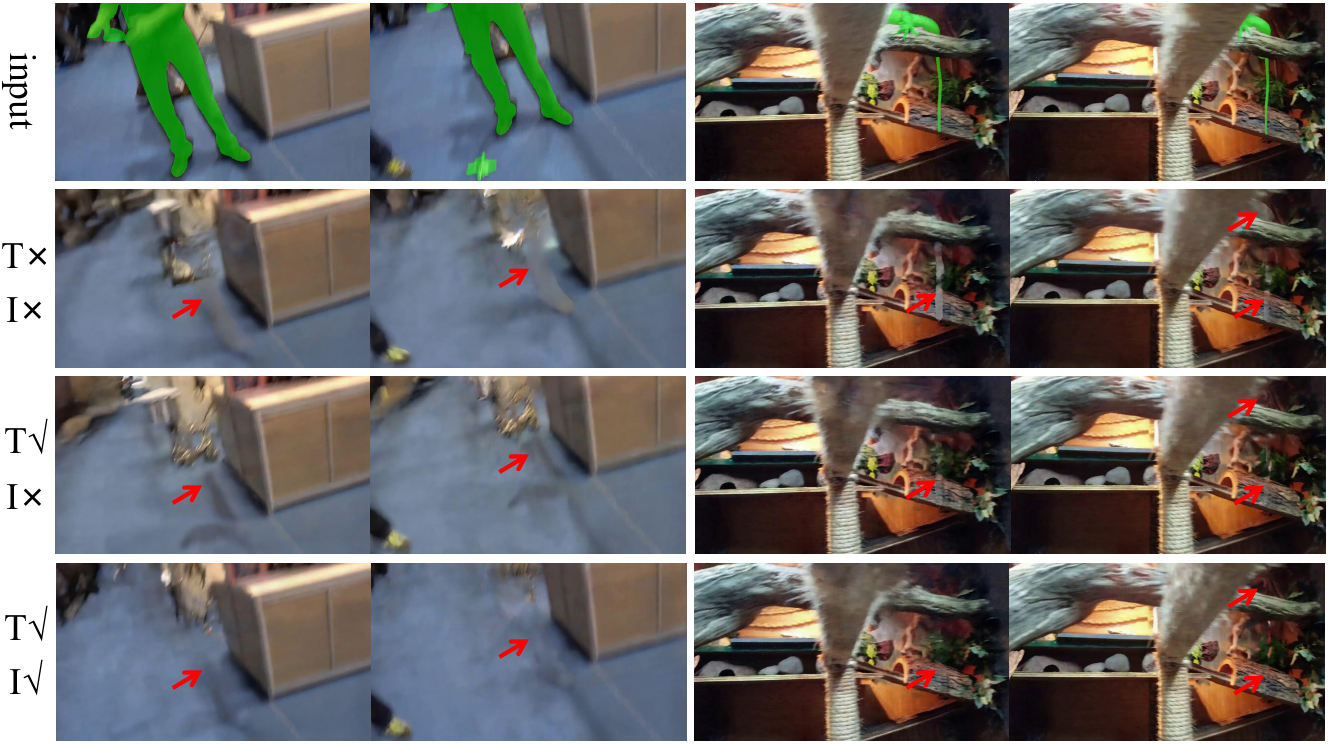}
    \caption{Effect of MUSE under abrupt-motion frames. MUSE improves removal even without additional training. ``T”/``I” denote Training/Inference, ``$\times$”/``$\checkmark$” indicate without/with MUSE.}
    \label{fig:stable_abrupt}
\end{figure}

\paragraph{\textbf{Stability for Abrupt-Motion.}}
As shown in \cref{fig:teaser}, existing diffusion-based video removal methods tend to fail on frames with abrupt motion, causing flicker or incomplete removal. The proposed MUSE strategy significantly mitigates this issue as shown in \cref{fig:stable_abrupt}. Without MUSE as in the second row, the removal failed. When integrating MUSE in the pipeline as shown in the fourth row, the new model can get stable removal for abrupt motion frames.

Moreover, as illustrated in the third row, MUSE serves as a \textit{training-free} and \textit{plug-and-play} enhancement that can be directly incorporated into existing methods to improve their robustness under abrupt motion. \textbf{This generalization capability is further validated in \cref{sec_supp:muse_others} in the supplementary material, }where MUSE consistently stabilizes results across multiple prior approaches.

\begin{figure}
    \centering
    \includegraphics[width=\linewidth]{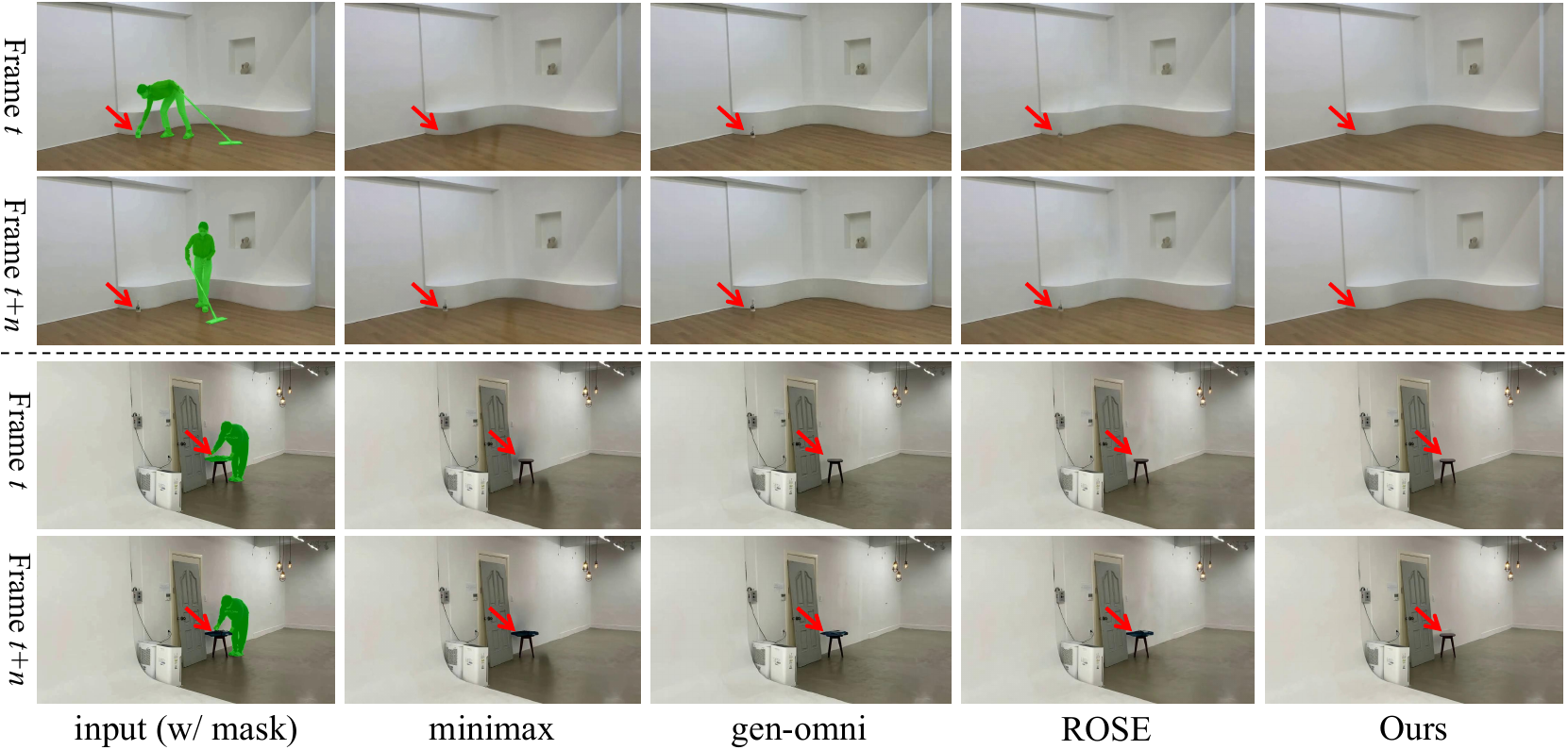}
    \caption{Robust removal under SAM2 failures. Existing methods miss unsegmented objects when SAM2 drops, while our \method still achieves temporally consistent removal.}
    \label{fig:stable_sam2}
\end{figure}

\paragraph{\textbf{Stability for SAM2 Segmentation.}}
To further evaluate the robustness of our method under realistic imperfect masks, we conduct experiments using segmentation masks generated by SAM2~\cite{ravi2024sam}. Specifically, we perform full segmentation on the first frame and propagate it through the video using SAM2 to obtain per-frame masks, which are then used for object removal.

As shown in \cref{fig:stable_sam2}, we compare several SOTA methods with ours under these real imperfect masks. When SAM2 occasionally fails to segment the object in certain frames, existing methods typically leave residual objects or incomplete erasures. In contrast, our method maintains stable removal performance, effectively handling the missing-mask cases caused by imperfect segmentation.

\paragraph{\textbf{Stability for Mask Degradation.}}
Our \method is also stable when dealing with imperfect segmentation masks. We simulate imperfect masks by randomly discarding mask frames at varying rates (from 0 to 50\%).

\begin{figure}[t]
  \centering
  \begin{subfigure}{0.49\linewidth}
      \centering
      \includegraphics[width=\linewidth]{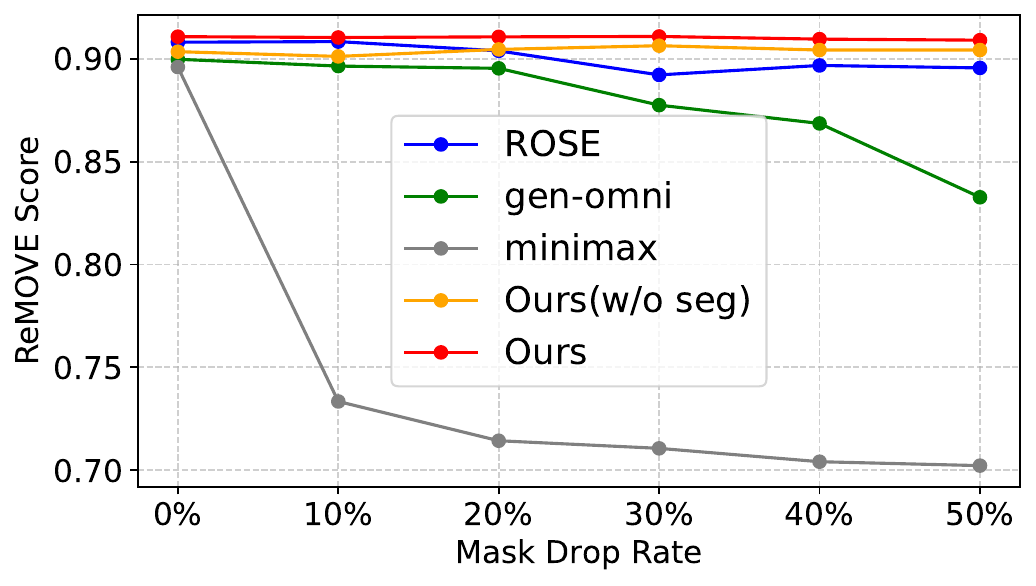}
      \caption{ROSE Bench}
      \label{fig:degrade_mask:rose}
  \end{subfigure}
  \begin{subfigure}{0.49\linewidth}
      \centering
      \includegraphics[width=\linewidth]{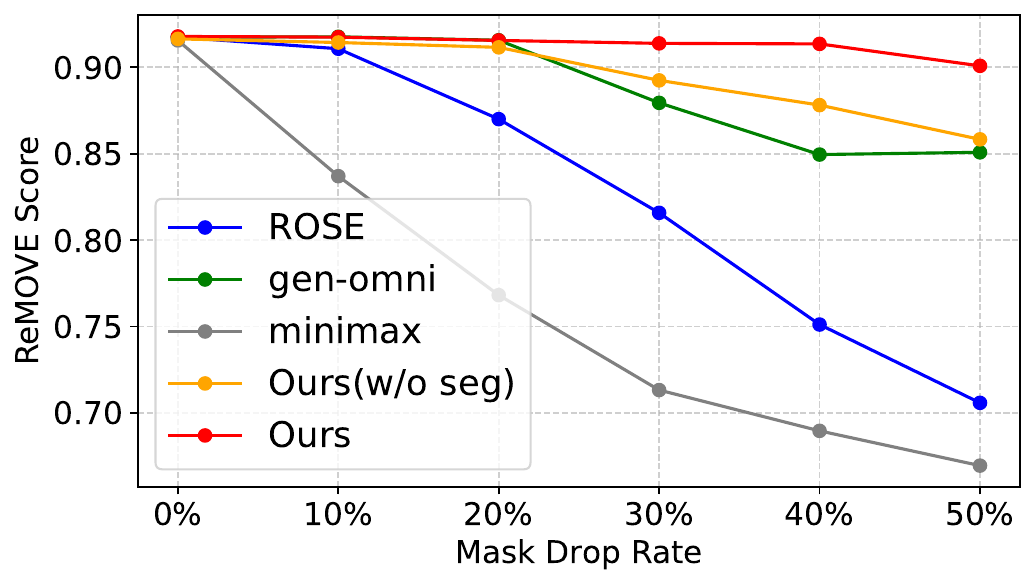}
      \caption{RORD-50}
      \label{fig:degrade_mask:rord}
  \end{subfigure}
  \caption{ReMOVE performance under mask drop. Our \method remains stable while existing methods collapse.}
  \label{fig:degrade_mask}
\end{figure}

\Cref{fig:degrade_mask} reports ReMOVE degradation across mask drop rate in the ROSE Bench and RORD-50 datasets. It indicates that as the mask degradation increases, existing methods show significant drops in ReMOVE score.
Benefiting from the mask degradation strategy, our method shows very limited performance degradation. When further equipped with DA-Seg, our \method gains the ability to implicitly reconstruct missing masks, consistently achieving the most stable and reliable removal results even with highly degraded masks across both datasets, demonstrating its robustness in handling imperfect masks. Notably, we observe that \method can, in some cases, accomplish reliable clip-wide removal given only a single-frame mask (see supplementary material \cref{sec_supp:single} for results).

\begin{table}
\caption{Ablation results of our strategies}
\label{tab:ablation}
\centering
\begin{tabular}{c|ccccc|cccc}
\toprule
Stage I & \multicolumn{5}{c|}{Stage II} & \multicolumn{4}{c}{ROSE Bench} \\
\midrule
bg\_train & Base & MUSE & MaskD & Seg & w-loss & PSNR$\uparrow$ & SSIM$\uparrow$ & LPIPS$\downarrow$ & Remove$\uparrow$ \\
\midrule
$\checkmark$ &  &  &  &  &  & 25.94 &  0.909 & 0.077 & 0.864   \\
 & $\checkmark$ &  &  &  &  &  27.33 & 0.918 & 0.066 & 0.901   \\
 & $\checkmark$ & $\checkmark$ &  &  &  & 27.87 & 0.921 & 0.061 & 0.902  \\
 & $\checkmark$ & $\checkmark$ & $\checkmark$ &  &  & 28.12 & 0.922 & 0.064 & 0.906  \\
 & $\checkmark$ & $\checkmark$ & $\checkmark$ & $\checkmark$ &  & 28.35 & 0.919 & 0.060 & 0.904   \\
 & $\checkmark$ & $\checkmark$ & $\checkmark$ & $\checkmark$ & $\checkmark$ & 28.68 & 0.927 & 0.052 & 0.907  \\
\midrule
$\checkmark$ & $\checkmark$ & $\checkmark$ & $\checkmark$ & $\checkmark$ & $\checkmark$ & 31.47 & 0.934 & 0.045 & 0.908 \\
\bottomrule
\end{tabular}
\end{table}

\subsection{Ablation Study}
\label{sec:exp_ablation}

We conduct ablation experiments on ROSE Bench to verify the effectiveness of the Stage~II strategies and the benefit of Stage~I pretraining. \Cref{tab:ablation} reports the results. \textit{Base} denotes training only on ROSE paired data without tricks. \textit{MaskD} applies the mask degradation strategy, \textit{Seg} adds the DA-Seg segmentation loss, \textit{w-loss} applies weighted diffusion loss, and \textit{bg\_train} denotes using the Stage~I background-pretrained model as initialization.

As shown in \cref{tab:ablation}, Stage~I pretraining already yields a strong baseline by providing realistic background and temporal priors. Each Stage~II component further improves performance. Combining all components with Stage~I initialization achieves the best overall performance, confirming the effectiveness of our curriculum design.

\section{Discussion}
\label{sec:conclusion}
We have presented \method, a stability-driven framework for video object removal under real-world imperfect conditions. By introducing \textbf{MUSE}, a temporal mask union strategy, we effectively mitigate localization failure in abrupt-motion frames. The proposed \textbf{DA-Seg} branch provides internal location priors and implicitly completes degraded masks, allowing robust removal when mask is imperfect. Furthermore, the \textbf{Curriculum Two-Stage Training} enables the model to first learn realistic background from unpaired data, and then refine removal fidelity and side-effect suppression under paired supervision.

Despite strong results, \method has several \textbf{limitations}: (1) under extreme sparsity (\eg, only a single-frame mask), \method may cause false/over-erasure; (2) constrained by existing datasets, it cannot fully remove all side effects.

%
%
\bibliographystyle{splncs04}
\bibliography{main}

\clearpage
\setcounter{page}{1}




\section{Supplementary Materials}

\subsection{Details of Background Data Construction}
\label{sec_supp:bgdata}

To support the background restoration warm-up in Stage~I, we construct a large-scale background-only video dataset from real-world open-source video collections, including OpenVid~\cite{nan2024openvid}, VideoGen~\cite{li2023videogen}, UltraVideo~\cite{xue2025ultravideo}, Inter4K~\cite{stergiou2022adapool}, and WISA-80K~\cite{wang2025wisa}. The goal is to automatically identify and remove video clips containing significant foreground objects, yielding a high-quality, unpaired background dataset. The pipeline consists of four steps:

\paragraph{(1) Quality Filtering.}
We first filter out videos with visual or semantic noise to ensure reliable training quality:
\begin{itemize}
    \item \textbf{Text Region Filtering:} We apply EasyOCR\footnote{\url{https://github.com/JaidedAI/EasyOCR}} to detect text areas in each frame. Clips containing large text regions such as subtitles, watermarks, embedded UI elements, or banners are discarded.
    \item \textbf{Aesthetic Score Filtering:} We compute aesthetic scores using LAION-Aesthetics\footnote{\url{https://github.com/barionleg/LAION-Aesthetics}} and discard samples below a threshold of 4.0.
    \item \textbf{Imaging Quality Filtering:} We employ MUSIQ~\cite{ke2021musiq} to evaluate blur, noise, and compression artifacts and remove clips with noticeable degradation.
\end{itemize}
This ensures that the remaining videos exhibit high clarity, minimal artifacts, and visually natural appearance.

\paragraph{(2) VLM-Based Scene Filtering and Balancing.}
We further apply SkyCaptioner-V1~\cite{chen2025skyreels}, a large-scale vision-language captioning model, to extract structured metadata including scene type, camera motion, and presence of salient subjects. Filtering and balancing are then performed as follows:
\begin{itemize}
    \item \textbf{Foreground Subject Screening:} If the generated description contains explicit dynamic subjects (\eg, ``a man walking", ``a car driving"), the clip is removed.
    \item \textbf{Scene Distribution Balancing:} We categorize remaining clips into semantic scene groups (\eg, natural landscape, streetscape, indoor environment). Sampling is performed to balance distribution across groups, preventing data bias toward a narrow background domain.
\end{itemize}

\paragraph{(3) Open-Vocabulary Instance Detection and Segmentation.}
To further validate the ``background purity" of filtered clips, we apply the open-vocabulary segmentation model DINOv3~\cite{simeoni2025dinov3}. Detected objects such as pedestrians, vehicles, and animals are measured by spatial area. Clips are discarded if any detected instance occupies more than 30\% of the frame area. This removes videos containing prominent or persistent foreground entities.

\paragraph{(4) Human Review and Final Quality Assurance.}
Finally, we manually review a subset of the remaining clips to remove subtle foreground presence, including mirror reflections, small human figures in the distance, or partial objects at the frame boundary. This step ensures strict dataset purity.

\paragraph{Final Dataset.}
Following the full pipeline, we obtain a high-quality background-only dataset containing approximately \textbf{49,000} video clips. The dataset spans diverse environments (urban, natural, indoor), camera motions, and lighting conditions, providing broad scene coverage and strong representational richness. This dataset enables stable and effective background prior learning in the Stage~I warm-up process.

\subsection{Details of the Random Mask Strategy}
\label{sec_supp:Rmask}
We propose a diversified random-mask generation strategy that procedurally synthesizes masks with rich motion characteristics, enhancing robustness and adaptability to diverse occlusion patterns. Concretely, we compose four spatial shapes with six temporal dynamics:
\begin{itemize}
  \item Spatial shapes
  \begin{itemize}
    \item Rectangle (fixed bounding box, bbox)
    \item Circle (fixed radius)
    \item Ellipse (fixed axis lengths)
    \item Full-frame mask (entire-frame occlusion)
  \end{itemize}
  \item Temporal dynamics
  \begin{itemize}
    \item Full-span mask: the same mask covers the entire video
    \item Interval mask: the mask appears only over a contiguous frame interval
    \item Per-frame random: each frame independently samples a random bbox (position/size)
    \item Per-frame jitter: small random offsets around an initial bbox each frame
    \item Constant-speed motion: bbox moves linearly at a fixed velocity
    \item Variable-speed motion: accelerated or non-linear trajectories (e.g., parabolic, S-shaped)
  \end{itemize}
\end{itemize}

Masks are sampled online during training, ensuring each clip encounters a different occlusion configuration. This improves generalization to occlusion location, shape, duration, and motion pattern. In particular, introducing motion masks (constant or variable speed) encourages learning of background reasoning in dynamic scenes, closely matching real-world erasure demands after object motion.

\subsection{RORD-50 Construction}
\label{sec_supp:rord50}
We construct the RORD-50 dataset to enable reliable evaluation under paired groundtruth conditions in real-world scenes. The process is described as follows:

We first select RORD cases that contain at least 30 consecutive frames and concatenate them into target videos requiring object removal. For each frame, we manually annotate the object mask. Although RORD provides pseudo groundtruth obtained via image inpainting models, these inpainted results often exhibit artifacts, incomplete removal, or temporal flickering, making them unsuitable as a reference for evaluating high-quality video removal.

Fortunately, the RORD scenes are captured with a static camera, and apart from the removed object, the scene background remains unchanged. Moreover, RORD provides clean background images without the target object. Therefore, for each selected video, we generate a groundtruth reference video by repeating its corresponding clean background image to match the original sequence length.

However, natural variations such as illumination changes or foliage movement may cause background inconsistencies between the original video and the clean reference. To ensure high-quality pairing, we compute the PSNR between the non-object regions of the target video and the constructed groundtruth sequence. Videos are ranked based on this background consistency score, and the top 50 highest-scoring sequences are retained.

The resulting dataset, \textbf{RORD-50}, contains 50 high-consistency paired video samples that provide reliable ground truth supervision for quantitative evaluation of video object removal.


\subsection{More Quantitative Results}
\label{sec_supp:more_quan_results}

\paragraph{\textbf{More Metrics.}}
We conduct a more comprehensive quantitative evaluation of existing diffusion-based state-of-the-art removal methods across multiple complementary dimensions, including perceptual similarity (LPIPS~\cite{zhang2018unreasonable}), temporal consistency (TC~\cite{zi2025minimax}), and context-aware removal quality (CFD~\cite{yu2025omnipaint}). In addition, on ROSE Bench and RORD-50, we report PSNR, SSIM, and LPIPS computed exclusively on non-removal (background) regions with respect to the ground truth, denoted as bgPSNR, bgSSIM, and bgLPIPS, respectively. The results are summarized in \cref{tab:cmp_more}. Overall, our method consistently achieves superior performance across most evaluation metrics.

\begin{table*}[t]
\caption{Extended quantitative evaluation on DAVIS, ROSE Bench, and RORD-50. The best performance is highlighted in \textbf{bold}, while the second-best is \underline{underlined}. All results are reproduced using the official implementations to ensure fairness.}
\label{tab:cmp_more}
\centering
\newcolumntype{I}{!{\vrule width 1.2pt}} 

\resizebox{\textwidth}{!}{
\begin{tabular}{cIc|c|cIc|c|c|c|c|cIc|c|c|c|c|c}
\toprule
\multicolumn{1}{cI}{} & \multicolumn{3}{cI}{DAVIS} & \multicolumn{6}{cI}{ROSE Bench} & \multicolumn{6}{c}{RORD-50} \\
\cline{2-16}
\multicolumn{1}{cI}{} &
bgLPIPS$\downarrow$ & TC$\uparrow$ & CFD$\downarrow$ & LPIPS$\downarrow$ & bgPSNR$\uparrow$ & bgSSIM$\uparrow$ & bgLPIPS$\downarrow$ & TC$\uparrow$ & CFD$\downarrow$ & LPIPS$\downarrow$ & bgPSNR$\uparrow$ & bgSSIM$\uparrow$ & bgLPIPS$\downarrow$ & TC$\uparrow$ & CFD$\downarrow$ \\
\midrule

DiffuEraser~\cite{li2025diffueraser} & \textbf{0.0211} & 0.9676 & \textbf{0.4453}
& 0.0885 & 28.5813 & 0.9214 & 0.0635 & 0.9870 & \textbf{0.3043}
& 0.0469 & 31.2622 & \underline{0.9506} & 0.0289 & 0.9878 & 0.3369 \\
gen-omni~\cite{lee2025generative} & 0.1048 & \textbf{0.9746} & 0.4608
& 0.1013 & 28.4689 & 0.9000 & 0.0791 & 0.9901 & \underline{0.3073}
& \underline{0.0442} & 31.7696 & 0.9308 & 0.0305 & \textbf{0.9973} & 0.3133 \\
minimax~\cite{zi2025minimax} & 0.0628 & \underline{0.9739} & 0.4484
& 0.0917 & 28.0605 & 0.9144 & 0.0678 & \textbf{0.9903} & 0.3134
& 0.0507 & 30.3690 & 0.9433 & 0.0318 & 0.9912 & \underline{0.3117} \\
ROSE~\cite{miao2025rose}  & 0.0631 & 0.9683 & 0.4805
& \underline{0.0770} & \underline{32.5895} & \underline{0.9292} & \underline{0.0417} & 0.9894 & 0.3486
& \textbf{0.0440} & \underline{32.2127} & 0.9361 & \textbf{0.0272} & 0.9915 & 0.3150 \\
Ours & \underline{0.0576} & \textbf{0.9746} & \underline{0.4464}
& \textbf{0.0451} & \textbf{33.2219} & \textbf{0.9443} & \textbf{0.0341} & \underline{0.9902} & 0.3336
& \textbf{0.0440} & \textbf{32.4764} & \textbf{0.9511} & \underline{0.0273} & \underline{0.9931} & \textbf{0.2855} \\

\bottomrule
\end{tabular}
}
\end{table*}

We observe that CFD exhibits somewhat counterintuitive behavior on ROSE Bench: models trained on the ROSE dataset (e.g., ROSE~\cite{miao2025rose} and Ours) obtain worse CFD scores. Therefore, we recommend interpreting CFD in conjunction with other metrics rather than in isolation.

\paragraph{\textbf{Category-wise Result on ROSE Bench.}}
\begin{table}[t]
\centering
\caption{Category-wise comparison on ROSE Bench. Best results are highlighted in \textbf{bold}. Overall, our method outperforms ROSE in the majority of categories.}
\label{tab:rose_cate}
\begin{tabular}{l|c|cccc}
\toprule
Category & Method & PSNR$\uparrow$ & SSIM$\uparrow$ & LPIPS$\downarrow$ & ReMove$\uparrow$ \\
\midrule
\multirow{2}{*}{Common}
 & ROSE & \textbf{36.60} & \textbf{0.952} & 0.041 & \textbf{0.859} \\
 & Ours & 36.37 & \textbf{0.952} & \textbf{0.022} & 0.853 \\
\midrule
\multirow{2}{*}{Shadow}
 & ROSE & \textbf{33.79} & 0.923 & 0.063 & 0.937 \\
 & Ours & 33.61 & \textbf{0.939} & \textbf{0.038} & \textbf{0.938} \\
\midrule
\multirow{2}{*}{Light}
 & ROSE & 30.07 & 0.921 & 0.086 & \textbf{0.900} \\
 & Ours & \textbf{31.16} & \textbf{0.932} & \textbf{0.041} & 0.896 \\
\midrule
\multirow{2}{*}{Reflection}
 & ROSE & 27.73 & \textbf{0.872} & 0.113 & 0.891 \\
 & Ours & \textbf{28.14} & 0.868 & \textbf{0.081} & \textbf{0.911} \\
\midrule
\multirow{2}{*}{Mirror}
 & ROSE & 28.35 & 0.938 & 0.088 & \textbf{0.933} \\
 & Ours & \textbf{29.29} & \textbf{0.961} & \textbf{0.042} & 0.923 \\
\midrule
\multirow{2}{*}{Translucent}
 & ROSE & \textbf{31.43} & 0.947 & 0.060 & 0.912 \\
 & Ours & 31.30 & \textbf{0.961} & \textbf{0.039} & \textbf{0.913} \\
\bottomrule
\end{tabular}
\end{table}

To further assess robustness across different object side effects, we follow the evaluation protocol of ROSE~\cite{miao2025rose} and report category-wise quantitative results on the six side-effect classes in ROSE Bench. As shown in \cref{tab:rose_cate}, our method effectively handles all categories and achieves superior performance over ROSE in most cases.

\subsection{Effectiveness of Stage I Background Data Pre-Training}
\label{sec_supp:bg_train}

\begin{table}[t]
\caption{Ablation study of our two-stage training scheme on ROSE Bench and RORD-50.}
\label{tab:two_stage}
\centering

\resizebox{\linewidth}{!}{%
\begin{tabular}{c|c|c|c|c|c|c|c|c}
\toprule
\multirow{2}{*}{} & \multicolumn{4}{c|}{ROSE Bench} & \multicolumn{4}{c}{RORD-50} \\
\cline{2-9} 
& PSNR$\uparrow$ & SSIM$\uparrow$ & LPIPS$\downarrow$ & Remove$\uparrow$ & PSNR$\uparrow$ & SSIM$\uparrow$ & LPIPS$\downarrow$ & Remove$\uparrow$ \\
\midrule
raw VACE    & 22.71 & 0.8802 & 0.1175 & 0.7154 & 19.21 & 0.8622 & 0.1395 & 0.6842 \\
copy-paste data & 25.78 & 0.9014 & \textbf{0.0770} & 0.8547 & 27.39 & 0.9310 & 0.0665 & 0.8420 \\ 
Background data & \textbf{25.94} & \textbf{0.9086} & {0.0773} & \textbf{0.8640} & \textbf{28.99} & \textbf{0.9350} & \textbf{0.0519} & \textbf{0.9135} \\
\midrule
Only Stage II & 28.68 & 0.9268 & 0.0522 & 0.9071 & 30.02 & 0.9357 & 0.0479 & 0.9163 \\
Stage I + II  & \textbf{31.47} & \textbf{0.9335} & \textbf{0.0451} & \textbf{0.9082} & \textbf{31.26} & \textbf{0.9378} & \textbf{0.0440} & \textbf{0.9179} \\
\bottomrule
\end{tabular}
}
\end{table}

\paragraph{\textbf{Stage I Pre-Training.}}

We first validate the effectiveness of Stage I training with background data. Specifically, we compare the original VACE model~\cite{jiang2025vace}, a variant finetuned using copy-paste data, and a variant finetuned using background videos. The copy-paste data are constructed based on VPData~\cite{bian2025videopainter}, where objects are randomly cropped from one video and pasted into another.

\begin{figure*}[t]
    \centering
    \includegraphics[width=\linewidth]{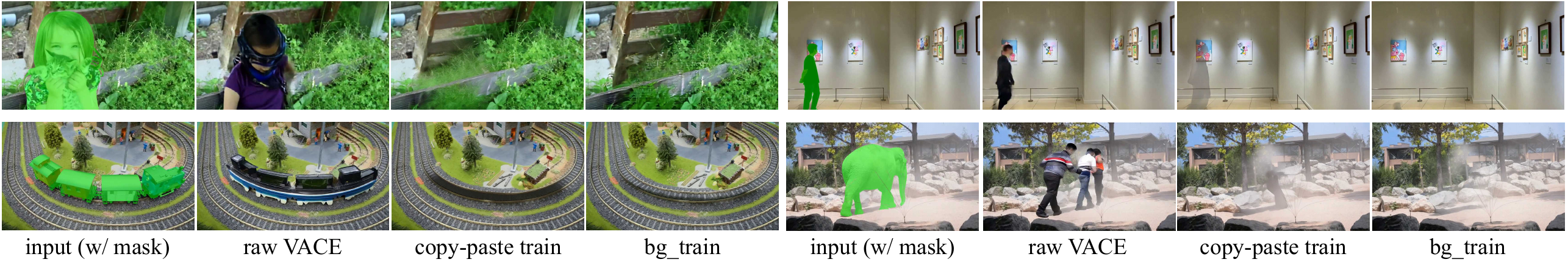}
    \caption{Effectiveness of Stage I pre-training. Training with background videos significantly improves removal quality and success rate.}
    \label{fig:bg_train}
\end{figure*}

As shown in the first three rows of \cref{tab:two_stage}, both finetuned variants achieve substantial improvements over the original VACE model across all metrics on ROSE Bench and RORD-50. Notably, the model trained with background videos consistently outperforms its copy-paste counterpart. The qualitative comparisons in \cref{fig:bg_train} further support this observation: the original VACE model tends to re-synthesize target objects, whereas both finetuned models effectively avoid this failure mode and achieve higher removal success rates. Moreover, training with background videos leads to better background completion quality and fewer filling artifacts than training with copy-paste data.

\begin{figure*}[t]
    \centering
    \includegraphics[width=\linewidth]{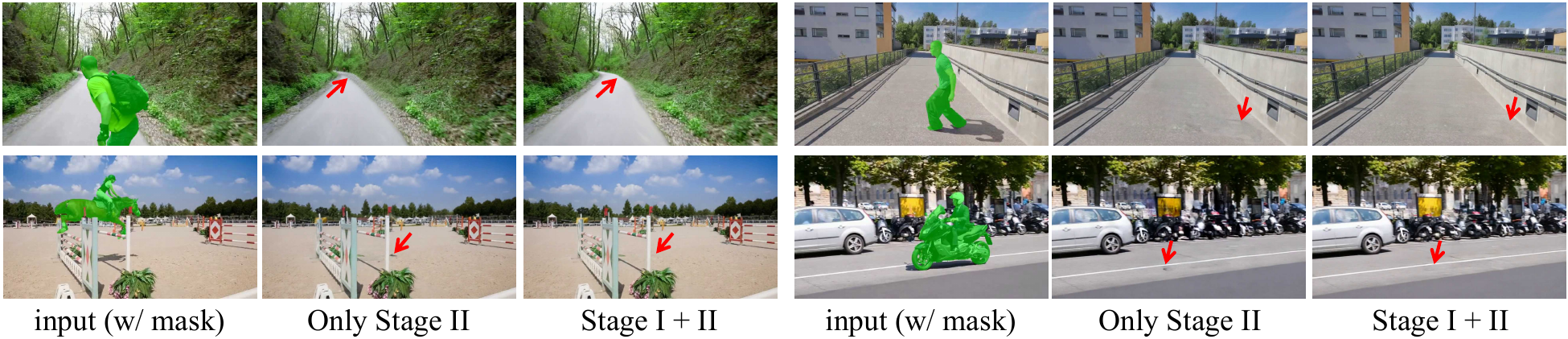}
    \caption{Comparison between Stage II–only training and the full two-stage training scheme. The complete two-stage training substantially improves background completion and shadow removal in real-world scenarios.}
    \label{fig:two_stage}
\end{figure*}

\paragraph{\textbf{Combining Stage I and Stage II.}}
We further investigate whether Stage I pre-training should be combined with Stage II refinement. The last two rows of \cref{tab:two_stage} compare models trained with Stage II only against those trained with the full two-stage scheme. The results show that the two-stage training consistently improves quantitative performance on both benchmarks. As visualized in \cref{fig:two_stage}, the full two-stage scheme yields more stable background completion and more reliable shadow removal in real-world scenarios, demonstrating the complementary benefits of Stage I pre-training and Stage II refinement.

\subsection{MUSE for Previous Models}
\label{sec_supp:muse_others}

\paragraph{\textbf{Qualitative Analysis.}}
We further evaluate the generality of MUSE on previous diffusion-based methods, including gen-omni~\cite{lee2025generative}, minimax~\cite{zi2025minimax}, and ROSE~\cite{miao2025rose}. MUSE is applied as a lightweight mask pre-processing step: masks are temporally grouped and unioned, and the resulting union mask is repeated to recover the original frame count.

\begin{figure*}[t]
    \centering
    \includegraphics[width=\linewidth]{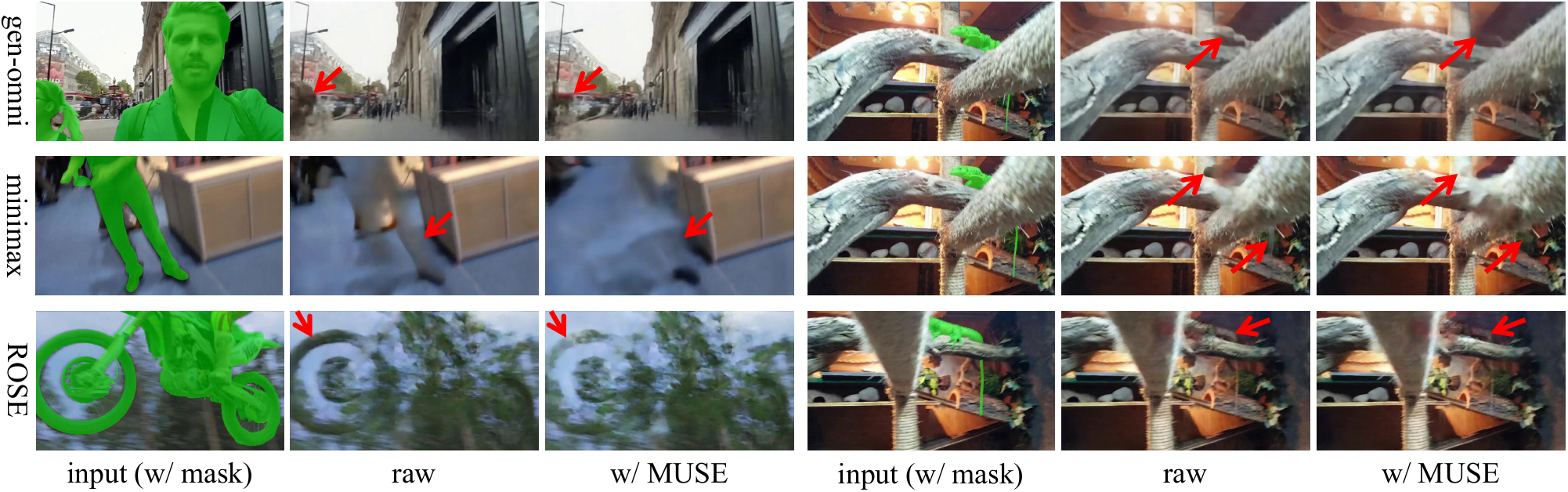}
    \caption{Effectiveness of \textbf{MUSE}. All methods suffer from missed removals or artifacts under abrupt motion, which is notably reduced after applying MUSE preprocessing.}
    \label{fig:muse_others}
\end{figure*}


As shown in \cref{fig:muse_others}, simply adding MUSE as a preprocessing step leads to clear improvements in abrupt-frame removal for all three methods. In some cases, residual artifacts at abrupt frames remain even with MUSE, which we attribute to the lack of joint integration of MUSE during model training. As demonstrated in ~\cref{sec:exp_stablity}, fully addressing this issue requires incorporating MUSE into the training process rather than applying it only at inference time.

\paragraph{\textbf{Quantitative Analysis.}}
Since abrupt frame displacement is rare in standard benchmarks, directly applying MUSE leads to only marginal metric changes on the original test sets. To better quantify its effect under such rare but critical conditions, we introduce a controlled temporal sparsification protocol. Specifically, we define a frame-skipping factor $k$, sampling one frame every $k$ frames to construct temporally sparser videos (and corresponding masks).

\begin{table*}[t]
\caption{Quantitative comparison with and without MUSE under different temporal compression ratios $k$ in ROSE Bench.}
\label{tab:skip_frame}
\centering
\resizebox{\linewidth}{!}{%
\begin{tabular}{c|c|c|c|c|c|c|c|c|c|c|c|c}
\toprule
\multicolumn{1}{c|}{} & \multicolumn{4}{c|}{$k=0$} & \multicolumn{4}{c|}{$k=2$} & \multicolumn{4}{c}{$k=4$} \\
\cline{2-13}
\multicolumn{1}{c|}{}
 & PSNR$\uparrow$ & SSIM$\uparrow$ & LPIPS$\downarrow$ & Remove$\uparrow$
 & PSNR$\uparrow$ & SSIM$\uparrow$ & LPIPS$\downarrow$ & Remove$\uparrow$
 & PSNR$\uparrow$ & SSIM$\uparrow$ & LPIPS$\downarrow$ & Remove$\uparrow$ \\
\midrule
gen-omni
& \bfseries{27.08} & 0.8831 & \bfseries{0.1013} & 0.8998
& 26.78 & \bfseries{0.8761} & \bfseries{0.1151} & 0.9011
& 26.46 & \bfseries{0.8698} & 0.1251 & 0.9014 \\

gen-omni w/ MUSE
& \bfseries{27.08} & \bfseries{0.8838} & 0.1014 & \bfseries{0.9003}
& \bfseries{26.80} & \bfseries{0.8761} & 0.1152 & \bfseries{0.9020}
& \bfseries{26.47} & 0.8697 & \bfseries{0.1250} & \bfseries{0.9027} \\

\midrule
minimax
& 26.30 & 0.8950 & \bfseries{0.0917} & \bfseries{0.8960}
& 26.00 & 0.8884 & \bfseries{0.0988} & 0.8968
& 25.89 & 0.8831 & \bfseries{0.1042} & 0.9001 \\

minimax w/ MUSE
& \bfseries{26.33} & \bfseries{0.8958} & 0.0918 & 0.8955
& \bfseries{26.05} & \bfseries{0.8885} & 0.0990 & \bfseries{0.8987}
& \bfseries{25.97} & \bfseries{0.8833} & \bfseries{0.1042} & \bfseries{0.9032} \\

\midrule
ROSE
& \bfseries{31.12} & \bfseries{0.9170} & 0.0770 & \bfseries{0.9081}
& 29.81 & 0.9019 & \bfseries{0.0671} & 0.9060
& 29.08 & 0.8920 & 0.0745 & 0.9016 \\

ROSE w/ MUSE
& \bfseries{31.12} & \bfseries{0.9170} & \bfseries{0.0665} & 0.9080
& \bfseries{29.85} & \bfseries{0.9020} & 0.0672 & \bfseries{0.9068}
& \bfseries{29.14} & \bfseries{0.8925} & \bfseries{0.0740} & \bfseries{0.9024} \\
\bottomrule
\end{tabular}
}
\end{table*}

We apply this protocol to ROSE Bench with $k=0,2,4$, where $k=0$ denotes the original test set. Quantitative results are reported in \cref{tab:skip_frame}. When $k=0$, the difference with and without MUSE is negligible. As $k$ increases, all evaluated methods consistently benefit from MUSE, with improvements becoming more pronounced under stronger temporal sparsification.

These results indicate that MUSE can be seamlessly integrated into existing models: it introduces no negative effects when abrupt motion is absent, yet yields clear gains when sudden frame transitions occur.

More broadly, MUSE highlights a common limitation of existing mask compression strategies in video inpainting, including direct temporal downsampling (e.g., VACE~\cite{jiang2025vace}, gen-omni~\cite{lee2025generative}), VAE-encoded masks (e.g., Minimax-Remover~\cite{zi2025minimax}), and folding time into channels (e.g., ROSE~\cite{miao2025rose}), which can all fail under abrupt motion. Addressing this issue more fundamentally remains an open problem.

\subsection{More Results of DA-Seg}
\label{sec_supp:seghead}

\paragraph{\textbf{Effectiveness of DA-Seg.}}
We further evaluate the effectiveness of the proposed DA-Seg on degraded-mask samples. For each case, we visualize the input masks, the predicted masks by DA-Seg, and the corresponding removal results. Since the predicted masks are downsampled, we use linear interpolation to restore them to the original temporal and spatial resolutions.

\begin{figure*}[t]
    \centering
    \includegraphics[width=\linewidth]{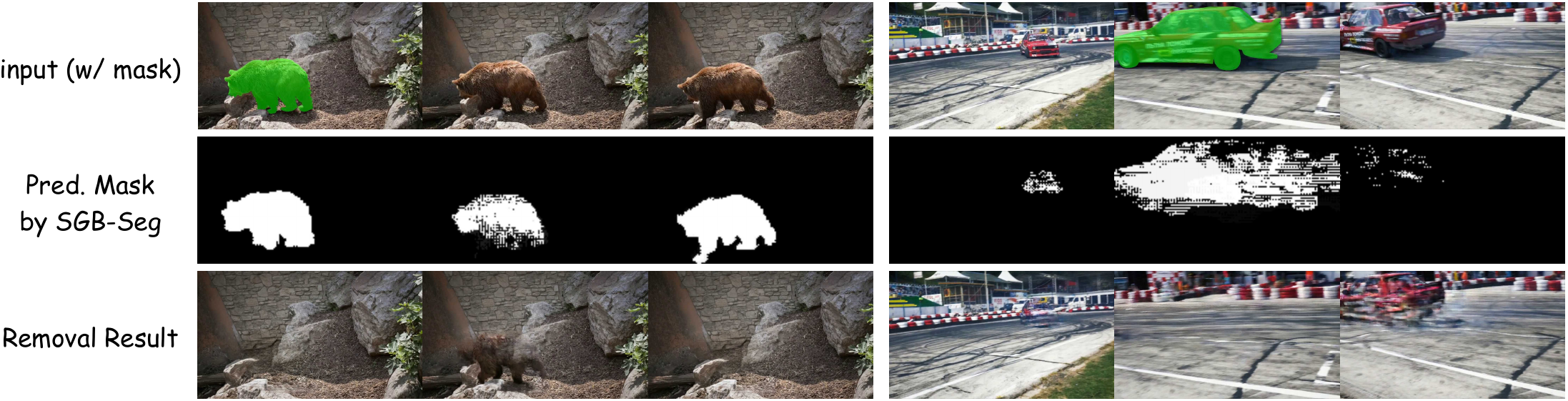}
    \caption{Effectiveness of \textbf{DA-Seg}. Accurate mask predictions lead to better removals, while broken masks cause degraded results.}
    \label{fig:seghead}
\end{figure*}

As shown in \cref{fig:seghead}, it can be observed that DA-Seg accurately completes defective masks, providing more reliable spatio-temporal guidance for object removal. Frames with accurate DA-Seg outputs yield cleaner and more consistent removal results. In some extreme cases when the input masks are severely degraded, if the DA-Seg prediction remains incomplete, the final removal quality is also negatively affected to some extent.

\paragraph{\textbf{Ablation of Segmentation Head Design.}}

\begin{table}[t]
\caption{Ablation study of our DA-Seg design on RORD-50.}
\label{tab:abl_seghead}
\centering

\resizebox{\linewidth}{!}{%
\begin{tabular}{c|c|c|c|c|c|c|c|c}
\toprule
\multirow{2}{*}{} & \multicolumn{4}{c|}{Mask Drop 0} & \multicolumn{4}{c}{Mask Drop 50\%} \\
\cline{2-9} 
& PSNR$\uparrow$ & SSIM$\uparrow$ & LPIPS$\downarrow$ & Remove$\uparrow$ & PSNR$\uparrow$ & SSIM$\uparrow$ & LPIPS$\downarrow$ & Remove$\uparrow$ \\
\hline
Vanilla Seg & 29.50 & \textbf{0.9360} & \textbf{0.0478} & 0.9170 & 25.07 & 0.9141 & 0.0835 & 0.8286  \\
DA-Seg & \textbf{30.02} & 0.9357 & 0.0479 & \textbf{0.9179} & \textbf{27.42} & \textbf{0.9220} & \textbf{0.0708} & \textbf{0.8599}  \\ 
\hline
\end{tabular}
}
\end{table}

We next conduct an ablation study on the design of the segmentation head. Specifically, we compare two variants: (i) a baseline head using standard LayerNorm (LN) without diffusion timestep conditioning, and (ii) our proposed head equipped with DA-AdaLN. \Cref{tab:abl_seghead} reports quantitative results on RORD-50 under two settings: using perfect ground-truth masks and randomly dropping 50\% of the mask frames. When perfect masks are provided, the performance difference between the two designs is marginal, indicating comparable capacity—this is expected, as the segmentation head plays a limited role when accurate masks are available. In contrast, under the more challenging setting with 50\% mask dropout, the DA-AdaLN-based head consistently outperforms the vanilla counterpart by a clear margin.

\begin{figure*}[t]
    \centering
    \includegraphics[width=\linewidth]{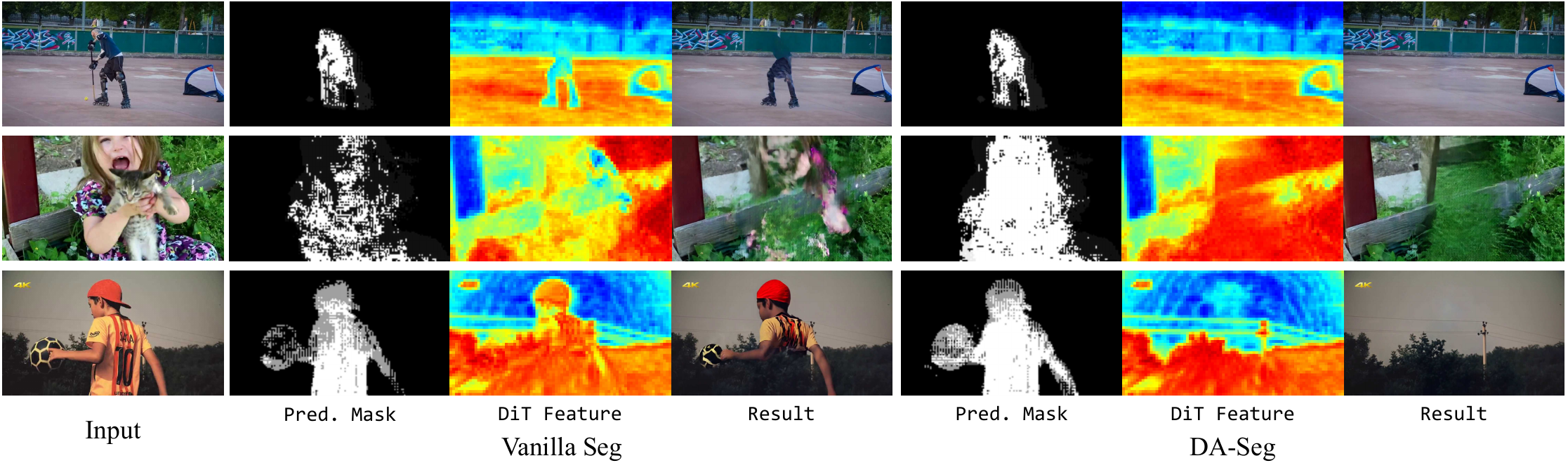}
    \caption{Ablation of segmentation head design. DA-Seg produces more accurate localization, indicating that the context block extracts more reliable control context for DiT. This enables the model to suppress the target object in latent features and achieve more stable removal under degraded mask guidance.}
    \label{fig:abl_seghead}
\end{figure*}

We further examine the segmentation outputs and corresponding removal results on DAVIS. As illustrated in \cref{fig:abl_seghead}, the DA-Seg head produces more accurate and cleaner localization than the vanilla segmentation head. This indicates that the context block guided by DA-AdaLN extracts more reliable control context for the DiT backbone. As a result, the DiT features are better aligned with the target region even when mask frames are missing, enabling the model to suppress the target object in the latent features and successfully remove it in the final output.

In contrast, the vanilla segmentation head lacks diffusion timestep conditioning and therefore introduces noticeably higher noise in its segmentation outputs. This noisy localization leads to inaccurate control signals, preventing the DiT features from consistently corresponding to the target region and ultimately causing removal failures.
Together with the quantitative results, these qualitative observations demonstrate that incorporating diffusion timestep conditioning via DA-AdaLN stabilizes context extraction and significantly improves removal performance under defective mask guidance.

\subsection{Single-frame mask-guided object removal}
\label{sec_supp:single}

\begin{figure*}[t]
    \centering
    \includegraphics[width=\linewidth]{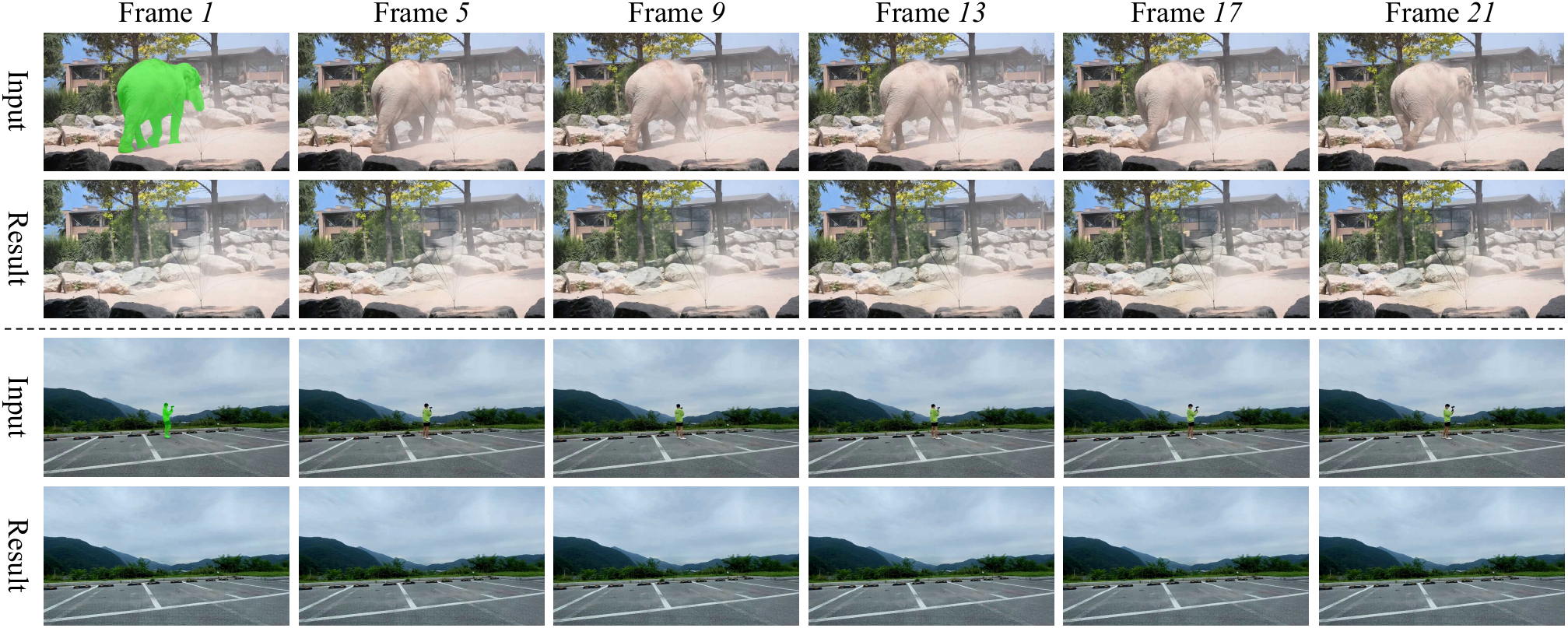}
    \caption{Results under single mask condition. In some cases, our \method can remove the target object even with only single mask.}
    \label{fig:single_frame}
\end{figure*}

MUSE prevents mask collapse introduced by temporal downsampling, while the combination of mask degradation and DA-Seg robustly handles imperfect mask guidance. These components are complementary and act synergistically: we observe that SVOR can already achieve clip-wide removal from a single-frame mask in a subset of scenarios (see \cref{fig:single_frame}). This capability is preliminary and not yet universal, but it motivates our next steps. Stabilizing the single-frame-mask regime would markedly improve speed and efficiency, simplify the pipeline, and enable broader real-world use.

\subsection{Details on GPT-based Evaluation}
\label{sec_supp:gpt_eval}

We additionally use \texttt{gpt-4o-2024-11-20} to automatically evaluate the perceptual quality of video removal results. For each video, we extract frames from the original video, the mask video, and the erased video. In the original frames, we overlay the mask region with a red highlight to clearly indicate the target removal area. Each evaluation input is therefore a pair: (Original Frame + Red Mask) and the corresponding Erased Frame.

GPT-4o is asked to score each frame from three dimensions: \textit{Target Removal Accuracy}, \textit{Visual Naturalness}, and \textit{Physical \& Detail Integrity}. The score of a frame is taken as the average of the three dimension scores. The final score of a video is computed by averaging the frame-level scores across all sampled frames.
This evaluation provides a perceptual measurement that complements quantitative metrics. The full prompt is provided in \cref{fig:gpt_prompts}.

\begin{figure*}[t]
\begin{tcolorbox}[title=Prompt for GPT-based Evaluation, colback=gray!20, colframe=black, fonttitle=\bfseries, sharp corners,fontupper=\tiny, fontlower=\tiny]
\textbf{Human:} You are a professional image rater evaluating object removal edits. You will be given two images:

- The first image is the original photo before editing. The red translucent area indicates the region to be removed, including all foreground objects that need to be eliminated.

- The second image is the result after object removal editing.

The objects removed may be humans, physical objects, text, or a combination of these.
Your task is to evaluate the object removal quality from **three perspectives**, each on a scale of 1 to 5.
For each score, provide a clear and specific explanation. Then, identify the removed object category and any new objects that were mistakenly generated.

---

**Target Removal Accuracy** (1–5): Did the system remove the correct object(s) completely, and only the intended object(s)?

1 – Nothing removed, or unrelated object removed.

2 – Target only partly removed, or wrong object/class removed, or new unintended object appears.

3 – Target mostly removed, but fragments remain, or extra objects were also removed.

4 – Only intended object(s) removed, but with minor collateral loss (\eg nearby detail lost, count incorrect).

5 – Perfect: all and only the specified object(s) removed; everything else preserved precisely.

**Visual Naturalness** (1–5): How natural and seamless does the edited image look?

1 – Severely broken (holes, artifacts, glitches, disfigured areas).

2 – Obvious erase marks, mismatched texture/color, or blurry/aliased patches.

3 – Acceptable, but has visible inconsistencies in lighting, texture, or resolution.

4 – Mostly natural with only minor artifacts visible when closely inspected.

5 – Seamless and photorealistic; indistinguishable from an unedited image.

**Physical \& Detail Integrity** (1–5): Is the structure, geometry, and realism of the scene preserved?

1 – Physically implausible result (\eg, floating limbs, warped objects, broken perspective).

2 – Major geometry or detail disruption; background structure lost or repeated unnaturally.

3 – Mostly consistent perspective and lighting; minor issues localized.

4 – Geometry and scene preserved; background logically filled in.

5 – Completely realistic and coherent, with well-restored structure, lighting, and fine details.

**Additional Tasks:**

- Identify the category/class of the object that was removed by comparing the two images.

- If any unreasonable or unintended new objects are generated in the result, list their class(es); if none, write `none`.

**Important:** The second and third scores must **NOT** exceed the first score.

---

\#\#\# Final Output Format (strictly follow this):

Target Removal Accuracy: $<$score 1–5$>$

Visual Naturalness: $<$score 1–5$>$

Physical \& Detail Integrity: $<$score 1–5$>$

Erase object class: $<$\eg, human, text, bottle, backpack, person+text$>$

Generate object class: $<$\eg, blur patch, ghost figure, floating object$>$, or `none'

$<$Image$>$ Image 1 $<$/Image$>$

$<$Image$>$ Image 2 $<$/Image$>$

\textbf{GPT:} ...
\end{tcolorbox}
\caption{Prompt for GPT-based evaluation}
\label{fig:gpt_prompts}
\end{figure*}

\end{document}